\documentclass[journal]{IEEEtran}
\usepackage{algorithm}
\usepackage[justification=centering]{caption}
\usepackage{algpseudocode}
\usepackage{fancyhdr}
\usepackage{textcomp}
\usepackage{amsmath}
\usepackage{subfigure}
\usepackage{amssymb,amsfonts}
\usepackage{graphicx}
\usepackage{textcomp}
\usepackage{xcolor}
\usepackage{multicol}
\usepackage{multirow}
\usepackage{blindtext}
\usepackage{adjustbox}
\algnewcommand{\Initialize}[1]{%
  \State \textbf{Initialize:}
\parbox[t]{.8\linewidth}{\raggedright #1}
}
\algnewcommand{\Goto}{\textbf{go to}}%

\def\BibTeX{{\rm B\kern-.05em{\sc i\kern-.025em b}\kern-.08em
    T\kern-.1667em\lower.7ex\hbox{E}\kern-.125emX}}

\usepackage{graphicx}
\usepackage{ulem}

\fancypagestyle{firstpage}{
    \fancyhf{} % 清空页眉页脚
    \lhead{This paper has been accepted by IEEE Transactions on Wireless Communications}
}

\begin{document}

\title{Intelligent Attacks and Defense Methods in Federated Learning-enabled Energy-Efficient Wireless Networks}

% \author{
%     \IEEEauthorblockN{Han Zhang\IEEEauthorrefmark{1}, Hao Zhou\IEEEauthorrefmark{1}, Medhat Elsayed\IEEEauthorrefmark{2},Majid Bavand\IEEEauthorrefmark{2}, Raimundas Gaigalas\IEEEauthorrefmark{2}, Yigit
%     Ozcan\IEEEauthorrefmark{2} and Melike Erol-Kantarci\IEEEauthorrefmark{1},\IEEEmembership{Senior Member, IEEE}}
    
%     \IEEEauthorblockA{\IEEEauthorrefmark{1} School of Electrical Engineering and Computer Science, University of Ottawa, Ottawa, Canada}
    
%     \IEEEauthorblockA{\IEEEauthorrefmark{2} Ericsson Inc., Ottawa, Canada}
    
%     \IEEEauthorblockA{\{hzhan363, hzhou098, melike.erolkantarci\}@uottawa.ca, \{medhat.elsayed;, majid.bavand, raimundas.gaigalas, yigit.ozcan\}@ericsson.com}
% }
\author{Han~Zhang,
        Hao~Zhou,
        Medhat~Elsayed,
        Majid~Bavand,
        Raimundas~Gaigalas,
        Yigit~Ozcan,
        and~Melike~Erol-Kantarci,~\IEEEmembership{Senior Member, IEEE}% <-this % stops a space
\thanks{Han Zhang, Hao Zhou and Melike Erol-Kantarci are with the School of Electrical Engineering and Computer Science, University of Ottawa, Ottawa, ON K1N 6N5, Canada (e-mail: hzhan363@uottawa.ca; hzhou098@uottawa.ca; melike.erolkantarci@uottawa.ca).}% <-this % stops a space
\thanks{Medhat Elsayed, Majid Bavand, Raimundas Gaigalas and Yigit Ozcan are with the Ericsson, Ottawa, K2K 2V6, Canada(e-mail:
medhat.elsayed@ericsson.com; majid.bavand@ericsson.com; raimundas.
gaigalas@ericsson.com; yigit.ozcan@ericsson.com)
}}% <-this % stops a space

%\author{\IEEEauthorblockN{Han Zhang, Hao Zhou, Medhat Elsayed, Majid Bavand, Raimundas Gaigalas and Melike Erol-Kantarci, \IEEEmembership{Senior Member, IEEE}}}

\markboth{ }%
{Shell \MakeLowercase{\textit{et al.}}: Bare Demo of IEEEtran.cls for IEEE Journals}

\maketitle

\thispagestyle{firstpage}
\begin{abstract}
%\red{Maybe we should design a new title with more specific techniques? what do you think? Similar idea as our magazine paper...and the lesson from the globecom paper. now, the title lacks some key technical words.}
%Federated learning (FL) is a promising technique for wireless networks due to its distributed learning features. However, due to the inherent vulnerabilities of FL, especially with non independent and identically distributed (non-IID) data, the attacks and defenses in FL have become an important topic and deserve more in-depth study.  
%In this work, we set up a federated deep reinforcement learning-based cell sleep control scenario and designed several intelligent attacks and defenses. Specifically, we designed a data poisoning attack, a generative adversarial network enhanced model poisoning attack, and a regularization-based model poisoning attack. In addition, we proposed two defense schemes, a similarity-based similarity-enabled defense and a knowledge distillation (KD) based defense. The simulation results demonstrate that the proposed attacks can degrade the network performance and lead to lower throughput and energy-efficiency. On the other hand, the proposed defense schemes can effectively protect the system from attacks. The system performance can be recovered to approximately 95\% of the performance under a secure system in the presence of the proposed KD-based defense. \red{I leave the old version here for your comparison.}
Federated learning (FL) is a promising technique for learning-based functions in wireless networks, thanks to its distributed implementation capability. On the other hand, distributed learning may increase the risk of exposure to malicious attacks where attacks on a local model may spread to other models by parameter exchange. Meanwhile, such attacks can be hard to detect due to the dynamic wireless environment, especially considering local models can be heterogeneous with non-independent and identically distributed (non-IID) data. 
Therefore, it is critical to evaluate the effect of malicious attacks and develop advanced defense techniques for FL-enabled wireless networks.  
In this work, we introduce a federated deep reinforcement learning-based cell sleep control scenario that enhances the energy efficiency of the network. We propose multiple intelligent attacks targeting the learning-based approach and we propose defense methods to mitigate such attacks. 
In particular, we have designed two attack models, generative adversarial network (GAN)-enhanced model poisoning attack and regularization-based model poisoning attack. As a counteraction, we have proposed two defense schemes, autoencoder-based defense, and knowledge distillation (KD)-enabled defense. The autoencoder-based defense method leverages an autoencoder to identify the malicious participants and only aggregate the parameters of benign local models during the global aggregation, while KD-based defense protects the model from attacks by controlling the knowledge transferred between the global model and local models. The simulation results demonstrate that the proposed attacks can degrade the network performance by 34\% and 77\%, and lead to lower throughput and energy efficiency. On the other hand, our proposed defense schemes can effectively protect the system from attacks. The system performance can be recovered to approximately 95\% of a secure system by using the proposed KD-based defense.

\end{abstract}

% \fi
\IEEEpeerreviewmaketitle

\section{Introduction}
%To meet increasing traffic demands and various service requirements,
%Machine learning (ML) has been widely applied to 5G and beyond 5G (B5G) networks, and 
Federated learning (FL) enables privacy-preserving collaborative learning in distributed systems and has been preferred in wireless networks due to its distributed nature as well as its benefits in terms of privacy and scalability.\cite{elsayed2019ai}. 
%FL is a promising technique for wireless networks 
Specifically, FL can utilize the on-device intelligence and train ML models in a decentralized manner while keeping the dataset in local servers \cite{zhang2023device}. 

While the effectiveness of FL-based network control applications has been demonstrated in previous studies \cite{zhang2022federated}, FL models are susceptible to attacks, which are rarely considered in these works \cite{bouacida2021vulnerabilities}. Therefore, it is a natural concern whether generic attacks against the FL models can impair the performance of FL-enabled wireless communication control applications, and in turn, affect wireless communication metrics such as throughput and energy efficiency. It is also an important concern to defend against these attacks. Considering the significant potential of FL, it is critical to study the security of FL-enabled wireless networks and evaluate the effect of attacks and defense methods especially in wireless network control application design.

In general, there are two types of attacks in FL, the poisoning attack \cite{tolpegin2020data} and the inference attack \cite{luo2021feature}. These two types of attacks have different objectives. They are usually discussed separately in existing research and are difficult to defend against simultaneously. In this work, we only focus on the poisoning attacks and corresponding defense methods. Among these attacks, a common way to attack FL is to manipulate the distributed participants and to make them submit malicious model parameters during the global aggregation. Such attacks on a local model can spread to other models by parameter exchange, which will degrade the overall system performance.
Considering the significant potential of FL in wireless networks and its vulnerability to attacks, it is critical to study the security of FL-enabled wireless networks.

%For instance, Sybil attacks \cite{patel2017review}, poisoning attacks \cite{sagduyu2019adversarial}, and jamming attacks \cite{pirayesh2022jamming} are all noticeable threats to the performance of wireless systems. Therefore, the security problem in AI-enabled wireless networks has become an important research topic and has attracted remarkable attention. \red{you consider ML as a background here, but the attack you mentioned, is it related to network intelligence? }

There have been many studies on such attacks in FL-enabled systems \cite{lyu2020threats}. 
%For instance, the data poisoning attacks are investigated in \cite{tolpegin2020data} and \cite{poi1sun2021data}, and \cite{fang2020local} which study model poisoning attacks against FL systems.
However, most existing attacks follow pre-determined rules and instructions and use a constant attack strategy, which makes them non-intelligent attacks. As a result, these attacks can be easily detected and defended by defender systems. By contrast, intelligent attacks usually can adapt their attack strategies to the changing environment. In particular, they can leverage ML techniques to gather, analyze, and respond to the information they collect from the interactions between the attacker and the environment. 
%For instance, in FL, by deploying a regularization between global and malicious model parameters, the malicious models can change the global model parameters, and then the attacks would be difficult to detect.
% \red{Still, can you give a more specific example here with an exact technology? It is less convincing without a specific example.}}
%In addition, they only consider FL systems without defenses or assume that the defense mechanism of the system is known to attackers, which is difficult to achieve in practice. Therefore, they can mimic to behaviours of benign participants and learn to attack the system in a more effective and stealthy way.
% \red{Can u give an example here to explain what is so-called intelligent attack?  }
There are two key reasons why we investigate intelligent attacks in this work. 
%First, we need to assess system vulnerabilities through more powerful attacks, which can help system operators better cope with potential outages. 
First, in terms of system evaluation, intelligent attacks can better identify system vulnerabilities. Evaluating the system performance under these attacks will help system operators cope with potential threats. 
%Second, more intelligent attacks allow us to make better comparisons between different defense mechanisms and develop improved ones.
Second, intelligent attacks can help testify to the capability of defense techniques since intelligent attacks provide more practical and stealthy approaches to undermine the system performance.

On the other hand, there are some existing studies on defense mechanisms for FL in wireless systems. %For instance, \cite{li2021lomar} proposes a defense method by comparing local model parameter distributions and \cite{wirelessdefense1kang2020reliable} performs defense with a consortium blockchain. 
Most of these defense techniques are evaluated  against non-intelligent poisoning attacks and simulated with independent and identically distributed (IID) data from standard data sets.
%. With the emergence of intelligent attackers, the previous defense mechanisms are likely to be ineffective. Second, the process of performing defense methods may introduce new vulnerabilities. Moreover, most defense-related studies only simulate independent and identically distributed (IID) data from standard data sets, which may be impractical in the real world \cite{panda2022sparsefed}. Specifically, these studies failed to consider the influence and difficulties posed by data heterogeneity of FL participants. 
We emphasize the issue of non-IID data in wireless network control scenarios, particularly for two reasons. First, due to the highly dynamic nature of the wireless environment and user behaviors, non-IID data is a more typical case in wireless scenarios. Second, data heterogeneity hinders the defenders' ability to identify malicious models as they may rely on identifying outlier behavior which can easily be confused with irregularities in non-malicious data. Therefore, there is a need to design more intelligent defense methods against intelligent attacks and to evaluate them in more realistic network application scenarios with non-IID data.

%and the security concerns with non-IID data are more challenging to deal with than with IID data. In non-IID cases, benign models with heterogeneous data distribution from the major data may be recognized as malicious participants\red{???}\noteblue{?what does this mean}. Instead, those true malicious models may become undetected due to the stealthy attack methods used by the attackers. Therefore, there is a need to design more intelligent defenses against intelligent attacks and to evaluate them in more realistic network application scenarios with non-IID data.

Inspired by these motivations, this work studies the intelligent attacks and defense methods in FL under wireless communication scenarios. Specifically, we first simulate a near-realistic wireless network environment and design a FL-based network control application. As a continuation of our previous work \cite{zhang2023distributed}, we choose sleep control as the wireless communication control application to evaluate the effect of attacks and defenses for two reasons. First, the cell sleep control problem is suitable to be solved by distributed cooperative training. 
%By changing the number of devices, traffic patterns, and average traffic loads of different cells, a non-IID scenario can be easily created. 
Second, the sleep control application is sensitive to the attacks. When the models are attacked, the sleep control application may control the base stations (BSs) to switch off in improper situations, so the attacks can directly lead to a significant decrease in throughput and energy efficiency. Improving network energy efficiency and reducing energy costs are critical goals for sustainable 5G and 6G networks. Combined with the sleep control scenario, the attacks and defense methods we propose not only affect the performance of the model but also demonstrate their practical impact in wireless communication scenarios.

The main contributions of this work are listed as follows:

1) We simulate a near-realistic wireless communication scenario and design a DRL-based sleep control application. This scenario provides a foundation for studying the security problems of FL-based control applications in wireless communication scenarios.

2) We propose two intelligent attack methods, a GAN-enhanced model poisoning attack, and a regularization-based model poisoning attack. These intelligent attacks can adapt to robust aggregation rules and bypass existing defense methods. Moreover, to the best of our knowledge, this is the first work that generates model parameters directly from GAN for model poisoning attacks. The simulations show that the proposed GAN-enhanced attacks and regularization attacks can lead to worse system performance than non-intelligent data poisoning attacks.

%We propose two intelligent attacks. \noteblue{Maybe we can remove the first sentence?} We state that the regular data poisoning attack is usually limited by the trade-off between the percentage of poisoned data and the level of difficulty in defense. Nevertheless, the GAN-enhanced model poisoning attack and the regularization-based model poisoning attack are more intelligent and stealthy methods. They can break such trade-off and bring more notable attack effects to Byzantine-robust FL systems. \red{Here you just need to summarize our former explanations of these two techniques. The trade-off things are not mentioned before. It can be weird.}
%Moreover, unlike existing research on GAN attacks, we do not exploit GAN to generate fake data for data attacks.\red{but why} 
%\red{I think we should be very careful about this section. Especially, which are our proposed methods, and which are existing works. I am thinking whether it makes sense to consider one of them as a baseline. Because proposing three attack models, actually weakens the contributions of each one.}\noteblue{Changed.}

3) In addition, we propose two smart defense strategies, autoencoder-based defense and KD-based defense. In autoencoder-based defense, we identify malicious participants according to the reconstruction errors of the model parameters through the autoencoder. By using an autoencoder, some of the impact of data heterogeneity on the defense can be reduced. In KD-based defense, we perform a distributed defense on the client side and combine KD with local verification, and this method can provide a guaranteed level of security.

It is worth mentioning that the main difference between these two defense methods is that autoencoder-based defense is performed at the global server and it defends against attacks by identifying malicious participants. It performs better for attack strategies where the attackers do not take any steps to make the attack more stealthy. However, autoencoder-based defense shows relatively worse performance in detecting intelligent attacks. KD-based defense is performed on the client side, and instead of finding malicious participants, it defends attacks by controlling the knowledge transferred between the global model and local models. Compared with autoencoder-based defense, KD-based defense shows higher effectiveness in intelligent and stealthy attacks. The proposed defense methods are tested with three attacks, a data poisoning attack baseline, and two of our proposed intelligent attacks. Both of these two defense methods can perform better than the Krum baseline, especially under intelligent attacks. We also analyze the capability of the KD-based method and derive an upper bound of the maximum effect of attacks with KD-based defense. 

The rest of the paper is organized as follows. Section II introduces related works. Section III shows the system model and problem formulations. Section IV introduces two intelligent attacks we proposed, and Section V presents the proposed two defense schemes. Finally, Section VI shows simulation settings and results, and Section VII concludes this work.

\begin{table}[]
\centering
\caption{Table of most frequently used notations}
\label{tab_notations}
\begin{tabular}{|c|c|c|c|}
\hline
Symbol &
  Description &
  Symbol &
  Description \\ \hline
$N$ &
  Number of SBSs &
  $b_n$ &
  Throughput \\
$M$ &
  Number of UEs &
  $P_0$ &
  \begin{tabular}[c]{@{}c@{}}Energy consumption\\ of MBS\end{tabular} \\
$R_n$ &
  The set of PRBs &
  $P_n$ &
  \begin{tabular}[c]{@{}c@{}}Energy consumption\\ of SBSs\end{tabular} \\
$B_r$ &
  Bandwidth &
  $a_n$ &
  Sleep mode \\
$\delta_{n}$ &
  Sleep Status &
  $\epsilon_m$ &
  Packet drop rate \\
$L_n$ &
  \begin{tabular}[c]{@{}c@{}}Traffic load \\ of SBSs\end{tabular} &
  $w$ &
  \begin{tabular}[c]{@{}c@{}}Coefficients to balance\\ different rewards\end{tabular} \\
$L_0$ &
  \begin{tabular}[c]{@{}c@{}}Traffic load \\ of MBS\end{tabular} &
  $\theta_n$ &
  \begin{tabular}[c]{@{}c@{}}Local model \\ parameters\end{tabular} \\
$\alpha$ &
  Learning rate &
  $\gamma$ &
  Discount factor \\
$\theta_G^{t+1}$ &
  \begin{tabular}[c]{@{}c@{}}Global model\\ parameters\end{tabular} &
  $L(.)$ &
  Loss function \\
$\theta_1$ &
  \begin{tabular}[c]{@{}c@{}}Encoder model \\ parameters\end{tabular} &
  $\theta_2$ &
  \begin{tabular}[c]{@{}c@{}}Decoder model \\ parameters\end{tabular} \\
$\theta_G$ &
  \begin{tabular}[c]{@{}c@{}}Generator model \\ parameters\end{tabular} &
  $\theta_D$ &
  \begin{tabular}[c]{@{}c@{}}Discriminator model \\ parameters\end{tabular} \\
$z_k$ &
  Generator input &
  $x_k$ &
  \begin{tabular}[c]{@{}c@{}}Real parameter\\  samples\end{tabular} \\
$k$ &
  GAN training set &
  $r$ &
  Reward \\
$a$ &
  Action &
  $s$ &
  State \\
$D$ &
  Distance &
  $q$ &
  Q-values of DQN \\ \hline
\end{tabular}
\end{table}

\section{Related Works}
There have been many studies that design attacks with malicious participants in FL. These attacks can be divided into two categories, data poisoning attacks \cite{tolpegin2020data}\cite{poi1sun2021data} and model poisoning attacks \cite{fang2020local}. Data poisoning attacks are usually performed by polluting the data set of the malicious participants involved in FL. For instance, \cite{tolpegin2020data} proposes an attack in a classification scenario by flipping the label of a source class to a target class. \cite{poi1sun2021data} proposes another efficient attack in FL by finding the implicit gradients of poisoned data and deriving optimal attack strategies.
On the other hand, \cite{fang2020local} formulates the model poisoning problem under byzantine aggregation rules into an optimization formulation and derives the malicious model parameters by solving the optimization. These studies mentioned above have provided considerable attack results, but most of them focus on classification problems and cannot be directly applied to other scenarios. Different from these works, we study attacks in deep reinforcement learning-enabled network control scenarios. Other than that, most of the existing works assume no defenders or known defense mechanisms while our proposed defenses can learn from dynamic system changes and adapt to evolving aggregation rules.
% there exist several limitations. These attacks are non-intelligent, which means that they usually follow pre-determined rules and instructions. They are not able to learn from dynamic system changes and adapt to evolving aggregation rules. In contrast, our proposed intelligent attacks can utilize ML techniques to learn adaptive attack policies. For instance, the GAN-enhanced model poisoning attack can use GAN to understand the parameter distribution of benign models. As a result, it can automatically change its attack strategy according to the change of benign models and aggregation rules.

It is worth mentioning that there has been some research on exploiting GANs for attacks in FL. For instance, in \cite{zhang2019poisoning}, a GAN-based poisoning attack is performed by generating fake training examples with GAN and giving flipped labels to these examples. Similarly, \cite{psychogyios2023gan} proposes a data poisoning attack by synthesizing a concatenated malicious data set with GAN. In existing works, the use of GAN in these studies is limited to generating fake data to compromise the model with data poisoning attacks. In this work, we directly generate fake models and perform model poisoning attacks with GAN.

Meanwhile, the defense methods against malicious participants in FL can be divided into two categories. The first method is to find malicious participants during global aggregation. For instance, in \cite{li2021detection},  the malicious participants in the FL system are identified with principal component analysis and K-means-based clustering. The second method is to add difficulties for attackers by changing the rules of global aggregation and model distribution. For example, \cite{itahara2021distillation} proposes an entropy reduction aggregation strategy to increase the robustness of FL by reducing the global logit entropy, and it has proved to be effective in defending against model poisoning attacks. Still, these defense methods mainly focus on classification problems and may not apply to reinforcement learning models in wireless communications. In addition, they usually defend against only the most common and straightforward attacks without simulating against the more intelligent ones.

The above defense schemes mainly focus on the cases of IID data, which assumes a perfect environment with identical data distributions. However, in real wireless communication environments, the number of devices, the distribution of BSs, and the arrival of traffic packets are usually non-identical.  
As a result, non-IID settings are more realistic and practical considerations, and it is critical to investigate the attacks and defense methods in a practical non-IID wireless environment. Some works focus on defense methods with non-IID data. In \cite{singh2023fair}, the non-IID problem is solved by distinguishing minority members from attackers based on micro-aggregation. \cite{ma2022shieldfl} uses the cosine similarity-based method and soft-bounded aggregation to defend against attacks in FL with non-IID data. The above-mentioned defense methods usually have requirements for implementation scenarios. They are more suitable to scenarios where there are a large number of participants and the attacks are non-intelligent. The effectiveness of the defense methods may decrease in the case of more powerful attacks or without a sufficient number of participants. In contrast, the defense method that we proposed applies to scenarios with fewer attacks and can provide a guaranteed level of security in the case of any type of attack, both intelligent attacks and non-intelligent attacks.

Other studies focus on possible attacks and defense methods when using FL in wireless communication scenarios. In \cite{zheng2022poisoning}, attacks specific to the wireless traffic prediction models are proposed. \cite{wirelessdefense1kang2020reliable} proposes a reliable FL scheme for mobile networks by using consortium block-chain as a trusted ledger to record and manage the reputation of participants. Existing attacks usually focus on supervised learning models, and their attack strategies may not apply to wireless network control applications that typically use reinforcement learning models. Meanwhile, existing defense strategies are mainly designed to defend against traditional attack methods and may have difficulty handling intelligent attacks. 

In summary, our proposed methods differ from existing works in the following aspects. Firstly, unlike most existing works that only test their algorithms with an IID data environment, we validate our algorithms in more realistic non-IID scenarios. Secondly, the proposed attack and defense methods are intelligent and can adapt their strategies to the changes in the FL system. Such intelligent attacks and defense can better evaluate the system vulnerability and protect the system from potential threats. Finally, the proposed defense methods do not have requirements for implementation scenarios and can provide a guaranteed level of security in the case of any type of attack, both intelligent attacks and non-intelligent attacks.

\section{System Model and Problem Formulation}
\begin{figure}[t]
\centering
\includegraphics[width=3.5in]{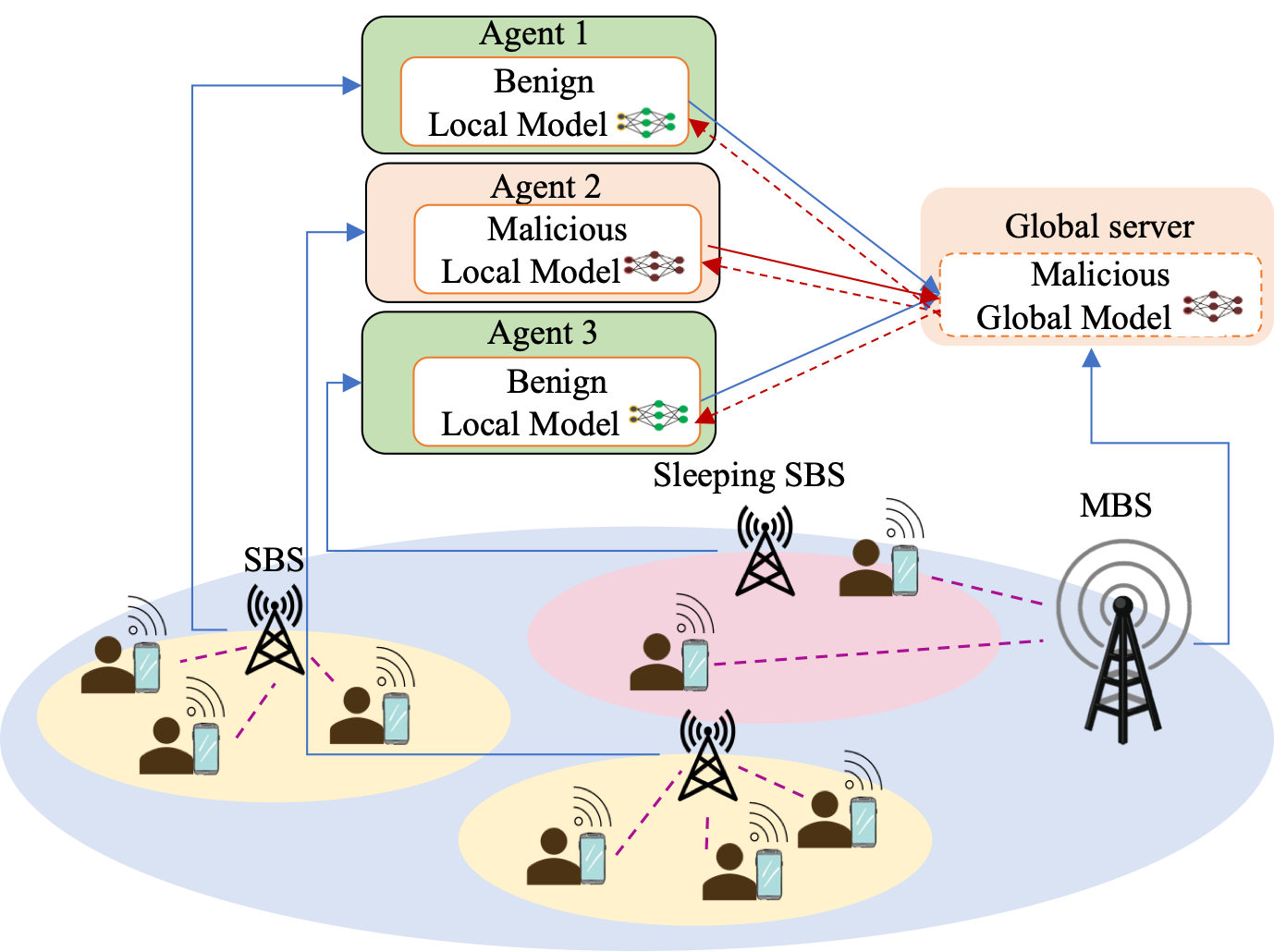}
\caption{Attacks in federated deep reinforcement learning-enabled cell sleep control scenario.}
\label{fig1}
\end{figure}

As shown in Fig. \ref{fig1}, we consider a heterogeneous network environment with one macro BS (MBS) and several small BSs (SBSs). To save energy, we perform dynamic sleep control on each SBS through its local model, while the MBS will stay active to ensure coverage and control of data services. On top of the deep learning models, the FL algorithm is performed to enable a privacy-preserving collaboration between all the local models. During the FL process, the MBS serves as the global server, aggregating models from SBSs and sending feedback to SBSs, which can accelerate local training. 

To investigate the intelligent attacks and defense methods in the proposed scenario, we assume that an attacker can compromise the cell sleep control application of more than one SBS \cite{zheng2022poisoning}. It can manipulate the local model parameters of compromised SBSs during the training process. 
As shown in Fig. \ref{fig1}, the sleep control application in the second SBS is compromised, so it provides malicious local model parameters to the global server during a single update in FL. The malicious model parameters will then be aggregated into the global model and fed to other local models, causing SBSs to choose non-optimal actions.

\subsection{System Model}
%To evaluate the influence of attacks and defenses, the system energy efficiency is regarded as a key indicator. 
We consider a downlink orthogonal frequency-division multiplexing cellular system and the link capacity between the $n^{th}$ SBS and the $m^{th}$ UE can be given as:
\begin{equation}
C_{n,m} = \sum_{r\in R_{n}} B_{r} log_{2}(1+SINR_{n,m,r}),\label{eq1}
\end{equation} 
where $R_{n}$ is the set of physical resource blocks (PRBs) allocated to the $n^{th}$ SBS. $B_{r}$ is the bandwidth of the $r^{th}$ PRB and $SINR_{n,m,r}$ is the signal interference noise ratio on the $r^{th}$ resource block. It can be formulated as:
\begin{equation}
SINR_{n,m,r} =\frac{\beta_{n,m,r} g_{n,m} P^T_{n}}{\underset{n'\in N, n'\neq n}{\sum}\ \underset{m'\in M_{n'}}{\sum} \beta_{n',m',r} g_{n',m} P_{n'}^T+B_{r}N_{0}},\label{eq2}
\end{equation} 
where $\beta_{n,m,r}$ denotes whether the $r^{th}$ resource block is allocated to the $m^{th}$ UE. $g_{n,m}$ is the channel gain of the transmission link. $P^T_{n}$ is the transmission power and $N_{0}$ is the noise power density. 

According to \cite{liu2015small}, the energy consumption model of the $n^{th}$ SBS can be formulated as:
\begin{equation}
    P_n = P^T_n + P^{AM}_n + P^{MP}_n + P^{FPGA}_n,\label{eq2-1}
\end{equation}
where $P^T_n$ is the transmission power. $P^{AM}_n$, $P^{MP}_n$, and $P^{FPGA}_n$ respectively denote the energy consumption of the power amplifier, microprocessor, and FPGA. The energy efficiency can then be calculated as:
\begin{equation}
EE = \frac{\sum_{n\in N} \delta_{n}b_n+b_0}{\sum_{n\in N}P_n + P_0},\label{eq3}
\end{equation}
where $\delta_{n}$ is a binary variable used to indicate the sleep status of the $n^{th}$ SBS. $b_n$ denotes the throughput of the $n^{th}$ SBS and $b_0$ denotes the throughput of the MBS. In our simulation, the throughput is determined by both the traffic pattern and the link capacity between BSs and UEs. $P_0$ and $P_n$ denotes the energy consumption of the MBS and the $n^{th}$ SBS.
%\noteblue{Both the throughput and power consumption $P_n$ are related to the sleep control variable $\delta_{n}$, so $\delta_{n}$ will dictate the final value of energy-efficiency $EE$. }

\subsection{Problem Formulation}
To save energy, we define three sleep modes for SBSs, active, sleep, and deep sleep. In active mode, SBSs are in full operation and consume the most energy. In sleep mode, the front end of the radio frequency transmitter is shut down to save about half of the energy consumption \cite{liu2015small}. Other hardware remains operational to ensure SBSs can be quickly woken up. In deep sleep mode, only the least amount of necessary energy will be consumed, but SBSs take longer to wake up.

The optimization objective of sleep control is to achieve high energy efficiency. The problem can be formulated as follows: 
\begin{align}
 \underset{a_n}{max}\ &EE, \label{eq4}\\
s.t.\ &(\ref{eq1})\ (\ref{eq2})\ (\ref{eq2-1})\ and\ (\ref{eq3})\tag{5a}
\\& a_n \in \{0,1,2\},\ \forall n \in N \tag{5b}\label{eq4a}
\\& \delta_n = \begin{cases}
1, &if\ a_n = 0,\\
0, &else.
\end{cases}\tag{5c}\label{eq4b}
\\& P_n = \begin{cases}
P_w, &if\ a_n = 0,\\
\gamma_1 P_w, &if\ a_n = 1,\\
\gamma_2 P_w,&if\ a_n = 2,\\
\end{cases}\tag{5d}\label{eq4c}
\end{align}
where $a_n$ denotes the sleeping modes of the $n^{th}$ SBS. $P_w$ denotes the energy consumption of the SBS in full operation. $\delta_n$ is a binary variable indicating whether the $n^{th}$ SBS is sleeping or not. $\gamma_1$ and $\gamma_2$ denote the ratio of energy consumption in the sleep mode and the deep sleep mode.

\subsection{Federated Learning for Cell Sleep Control}
We consider each SBS as a distributed agent and set up a FL framework. It aims to make distributed agents cooperatively train a shared global model while keeping local data private. This cooperative training approach allows SBSs to train a generalized DQN model with limited local data that can adapt to the dynamic changes in the environment. The Markov decision process (MDP) of each local DQN can be given as:
\begin{itemize}
    \item State: The state of sleep control is composed of four components, which are the sleeping status of the SBS $\delta_n$, the traffic load of the SBS and MBS in the past 5 transmission time intervals, and the current throughput of the SBS. 
    It can be formulated as:
    \begin{equation}
     s_n = \{\delta_n, L_n, L_0, b_n\},\forall n \in N.\label{eq6}
     \end{equation}
     Among these components, $L_n$ denotes the traffic load of the SBS in the past 5 transmission time intervals. It indicates the traffic pattern of the SBS. $L_0$ is the traffic load of the MBS in the past 5 transmission time intervals. It indicates the ability of the MBS to perform transmission tasks when the SBS is in sleep mode. $b_n$ is the current throughput of the SBS. The SBS can change the sleep mode accordingly to deal with the incoming traffic.
     
     \item Action: The action is to choose an optimal sleeping mode for the SBS, which can be formulated as:
     \begin{equation}
     a_n = \{0,1,2\}.\forall n \in N.\label{eq7}
     \end{equation}
     \item Reward: The reward function is defined as a combination of throughput, packet loss rate, and energy cost, which can be given as:
     \begin{equation}
      r_n = \omega_1 b_n - \omega_2\epsilon_n  - \omega_3 P_n,\forall n \in N,\label{eq8}
     \end{equation}
     where $\epsilon_m$ is the packet loss rate of the $m^{th}$ UE. $\omega_1$, $\omega_2$ and $\omega_3$ are the coefficients to balance different rewards.
\end{itemize}

 In each FL cycle, the local models at SBSs will first perform local training according to local experience. The local training process can be given as:
\begin{equation}
\begin{split}
\theta_n^{t+1} &= \theta_n^t + \alpha[r_n^t + \gamma \mathop{max}\limits_{a}Q(s_n^{t+1},a;\theta_n^t) \\ & -Q(s_n^t,a_n^t;\theta_n^t)]\nabla Q(s_n^t,a_n^t;\theta_n^t),\label{eqrl}
\end{split}
\end{equation}
where $\theta_n$ denotes the local model parameters of the $n_{th}$ SBS, $\alpha$ denotes the learning rate and $\gamma$ denotes the discount factor. $Q(s_n^t,a_n^t;\theta_n^t)$ denotes the long-term expected reward of the $n_{th}$ SBS choosing the action $a_n^t$ under the state $s_n^t$.

After that, local parameters will be uploaded to the global server. A global model will then be generated by aggregating the local parameters. Since we set the same size of memory buffer for all the SBSs, the aggregation weight of each local model is the same and equal to $\frac{1}{N}$ in this scenario. The global aggregation can be defined as:
\begin{equation}
\theta_G^{t+1} = \sum_{n=1}^{N}\frac{1}{N}\theta_n^{t+1},\label{eq_ag}
\end{equation}
where $\theta_G^{t+1}$ denotes the global model parameters and $\theta_n^{t+1}$ denotes the local model parameters of the $n^{th}$ SBS.

\section{Intelligent Attacks: GAN-enhanced and Regularization-based Attacks.}
This section introduces two proposed intelligent attacks in the federated deep reinforcement learning-based sleep control scenario, the GAN-enhanced model poisoning attack, and the regularization-based model poisoning attack. 

\subsection{GAN-enhanced Model Poisoning Attack}

In regular data poisoning attacks, attackers destruct model parameters by polluting training data. These attacks can successfully attack the wireless system when there are enough compromised SBSs. However, one downside is that the distribution of malicious model parameters may differ from benign model parameters. Therefore, malicious local models can be easily detected and defended by checking the similarity of all the local models with Byzantine-robust aggregation rules\cite{blanchard2017machine}. In contrast, a more stealthy and effective attack strategy is to exploit the strength of the GAN technique and generate malicious fake model parameters with similar distributions to benign model parameters.

\begin{figure*}[!t]
\centering
\includegraphics[width=6in]{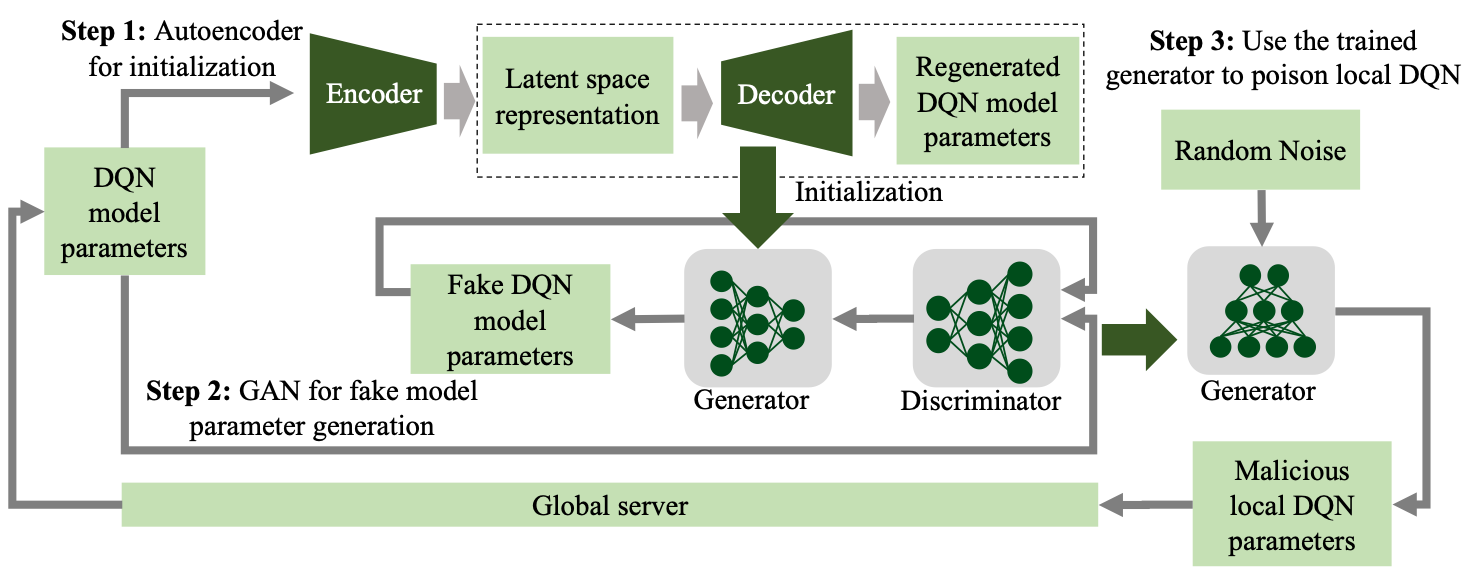}
\caption{The procedure of GAN-enhanced model poisoning attack.}
\label{fig2}

\end{figure*}

Fig. \ref{fig2} illustrates our proposed GAN-enhanced model poisoning attack. The whole procedure can be divided into training and attack phases. The training phase includes autoencoder-based initialization and GAN training for fake model parameter generation. The attack phase is to use the trained generator to poison local DQNs. 

One of the main difficulties of generating fake model parameters using GAN is the high dimensionality of the model parameters and the limited number of training samples that make training more challenging. Since the only source of real model parameter samples for the compromised SBSs is through the global model distribution, the compromised SBSs can only obtain one training sample during each iteration. That means the GAN of attackers only has limited training samples.
To handle such difficulty, GAN is only used to generate parameters of the bottom layer of the local model instead of the whole model.

In the first iterations, the compromised SBSs remain in regular training and accumulate real model parameter samples until sufficient training samples are available. Then they are trained on the autoencoder with collected training samples. An autoencoder is a self-supervised learning algorithm designed to enable neural networks to learn to compress and reconstruct high-dimensional data by using the same inputs and outputs. The problem can be formulated as:
\begin{align}
    \underset{\theta_1,\theta_2}{\operatorname{argmin}}\ L(x_{input},f_2(f_1(x_{input};\theta_1);\theta_2)),\label{eq_GAN1}
\end{align}
where $L$ denotes the loss function. $f_1$ and $f_2$ denote the encoder and the decoder. $x_{input}$ denotes the input data. $\theta_1$ and $\theta_2$ denote the model parameters of the encoder and the decoder. After training, the decoder can reconstruct the high-dimensional data from the low-dimensional compressed representations. Therefore, it could be an initialization to the generator to ensure the GAN will start training from a good initial solution \cite{mariani2018bagan}.

The core idea of GAN is to use two neural networks, a generator network, and a discriminator network, to perform adversarial training. The task of the generator network is to generate fake model parameters that the discriminator could not identify. Therefore, the loss of the generator network over a training set $K$ can be calculated as:
\begin{align}
    L_G(\theta_G) = \frac{1}{|K|}\sum_{k\in K}(\log(1-D(G(z_k;\theta_G);\theta_D))),\label{eq_GAN2}
\end{align}
where $\theta_G$ and $\theta_D$ denote the model parameters of the generator and the discriminator. $z_k$ denotes the input of the generator, which is randomly generated noise. On the other hand, the task of the discriminator network is to identify fake model parameters from real ones. The loss of the discriminator network can be calculated as:
\begin{align}
    L_D(\theta_D) = \frac{1}{|K|}\sum_{k\in K}(\log D(x_k;\theta_D)+\\ \log(1-D(G(z_k;\theta_G);\theta_D))),\nonumber
\end{align}
where $x_k$ denotes the real model parameter samples.

After training the GAN, we enter the attack phase and use the generator to generate malicious local model parameters. In the attack phase, the random noise generation pattern differs from the input noise in the training phase. This ensures the generator output will have a similar distribution but contain different information from the other benign local parameters. The output will then replace the parameters of the global model received from the last model aggregation and submit the new malicious local model to the global server. 
The GAN-enhanced model poisoning attack approach is summarized in Algorithm \ref{alg:GAN}. $\theta_{Local}(L_{out})$ denotes the parameters of the output layer of the local model. By generating fake model parameters with GAN, the malicious model parameters will share a similar distribution with benign model parameters. Therefore, it is difficult for the defenders to defend against GAN-enhanced model poisoning attacks by comparing the similarity between local parameters. This makes malicious models more stealthy and more resistant to detection.

\begin{algorithm}[t]
\caption{GAN-enhanced model poisoning attack}\label{alg:GAN}
\begin{algorithmic}[1]
    \Initialize{$\theta_{Local}$ of each local model, $s^{t-1}$ and $a^t$}
    \While {$TTI < T^{total}$}
        \For{benign SBSs}
            \State Observe local state.
            \State $S^{t}_n \gets \{\delta_n, L_n, L_0, d_n, b_n\}$.
            \State Get immediate reward $r^t_n$.
            \State Record experience $\{s^{t-1}_n,s^{t}_n,a^t_n,r^t_n\}$.
            \State Sample a mini-batch from experience buffer
            \State Update local Q network and get $\theta_{Local}$
            \State Choose actions by $a^t_n \gets \underset{a^t_n}{max}Q(s,a;\theta_{Local})$.
            \State Execute action $a^t_n$. 
        \EndFor
        \For{malicious SBSs}
            \If {$n(GAN samples) > batch\ size$}
                \State Generate noise for training $z_k^T \sim N(\mu_1,\sigma_1^2)$
                \State Update $
                \theta_G$ with according to Equation (\ref{eq_GAN1}) and Equation (\ref{eq_GAN2})
                \State Generate noise for model inference 
                \State $z_k^R \sim N(\mu_2,\sigma_2^2)$
                \State Get fake local model parameters
                \State $\theta_{Local}(L_{out}) = G(z_k^R;\theta_G)$
            \Else 
                \State $GAN samples\gets \theta_{Global}(L_{out})$
            \EndIf
        \EndFor
        \State Update global model according to Equation (\ref{eq_ag})
        \For{each UE} 
            \State Update local models $\theta_{Local}=\theta_{Global}$
        \EndFor
        \State $TTI \gets TTI + 1$
    \EndWhile
\end{algorithmic}
\end{algorithm}

\subsection{Regularization-based Model Poisoning Attack}
Another promising and more concise approach to attack the FL system is to directly change the direction of gradient descent of malicious local models by changing the loss function of the compromised SBSs. On the other hand, we expect the attack to be stealthy enough and be able to still successfully attack the system under byzantine-robust FL aggregation rules \cite{blanchard2017machine}. 
Therefore, there are two criteria to follow while designing the loss function of malicious models. Firstly, the output of malicious models is expected to be different from the benign models. Secondly, the model parameters of malicious models are expected to be close to the benign models \cite{fang2020local}\cite{bhagoji2019analyzing}. 

The first criterion can be achieved by maximizing the loss function instead of minimizing it during local training. The second criterion can be achieved by adding a regularization term to the loss function to limit the distance between the malicious and benign model parameters. Therefore, the loss function can be re-formulated as:
\begin{align}
    L(\theta_{Local}^t) = & (r^t+\gamma\max_a Q(s^{t+1},a;\theta_{Local}^t)+ \label{eq_regu1}\\ &Q(s^t,a^t;\theta_{Local}^t)))^2 +\omega||\theta_{Local}^t-\theta_{Global}^t||_2,\nonumber
\end{align}
where $\theta_{Local}$ and $\theta_{Global}$ denote the local and global parameters. $r$ is the reward. $\gamma$ is the discount factor of reinforcement learning. $||.||$ denotes the L2-Squared distance. $\omega$ is a coefficient used to balance two sources of the loss. The local update of malicious models can be formulated as:
\begin{align}
    \theta_{Local}^{t+1} &= \theta_{Local}^{t} - 2\alpha(r+\gamma\max_a Q(s^{t+1},a;\theta_{Local}^{t})\label{eq_regu2}\\&+Q(s^{t},a^{t};\theta_{Local}^{t}))\nabla Q(s^{t},a^{t};\theta_{Local}^{t})\nonumber \\ &+Q(s^{t},a^{t};\theta_{Local}^{t}))-2\omega\alpha(\theta_{Local}^{t}-\theta_{Global}^{t}),\nonumber 
\end{align}
where $\alpha$ denotes the learning rate. $\nabla(.)$ denotes the differential value.

To minimize the first term of the loss function, the output Q values of the network will represent the complementary values of the long-term reward for given actions in given states. By choosing the action with the maximum Q value, the malicious model will select the action minimizing the long-term reward. In this way, the compromised SBSs will generate malicious model parameters during local training with a modified loss function, leading to attacks on the system through malicious local models.

\subsection{ Baseline Method: Data Poisoning Attack}
To make a comparison with two proposed intelligent attacks, we use a non-intelligent data poisoning attack as a baseline. It is implemented by poisoning some of the MDP data in the local memory buffer of the compromised SBSs so that malicious local models are trained in the wrong directions \cite{tolpegin2020data}. According to the settings of the given scenario, high reward values are usually scarce but contain critical information. Therefore, in the sleep control application, the data poisoning attacker can select training records with high reward values from the local memory buffer of the compromised SBSs and poison the data by turning positive rewards into negative rewards. In this way, by poisoning only a small fraction of the data, the attacker can make malicious local models learn and share the opposite experience to the global model and benign local models.

\subsection{Computational complexity and overhead analysis}
In this subsection, we analyze the computational complexity and the overhead of two proposed attacks. For the GAN-enhanced model poisoning attack, the main computation comes from the training and inference of GAN models. The complexity of the GAN model is proportional to the model layers and the number of neurons of both the generator and the discriminator. So it can be given as $O(\sum_{i=1}^{I_G}N^G_{i-1}N^G_{i}+\sum_{D_i=1}^{I_D}N^D_{i-1}N^D_{i})$, where $N^G_{i}$ denotes the number of neurons of the generator at the $i^{th}$ layer. $N^D_{i}$ denotes the number of neurons of the discriminator at the $i^{th}$ layer. $I_G$ and $I_D$ denote the number of layers of the generator and the discriminator.

Regularization-based model poisoning attack brings almost no extra computations. The only added step is to calculate the Euclidean norm between the local model parameters and the global model parameters and the computational complexity is proportional to the dimension of model parameters $len(\theta)$.

Both of the attacks do not introduce additional overhead to the system. The same as FedAvg, the only information exchanged between the global server and the local servers are the local model parameters and the global model parameters.

\section{Intelligent Defense Methods: Autoencoder and KD-based Defense }
This section introduces two proposed defense schemes in federated
deep reinforcement learning-based sleep control scenario, an autoencoder-based two-step defense, and a KD-based defense. The first defense scheme is performed at the global server and based on the similarity of local model parameters. The second defense scheme is mainly performed on the client side as a combination of heterogeneous FL and local data-based validation.

\begin{figure}[!t]
\centering
\includegraphics[width=3.5in]{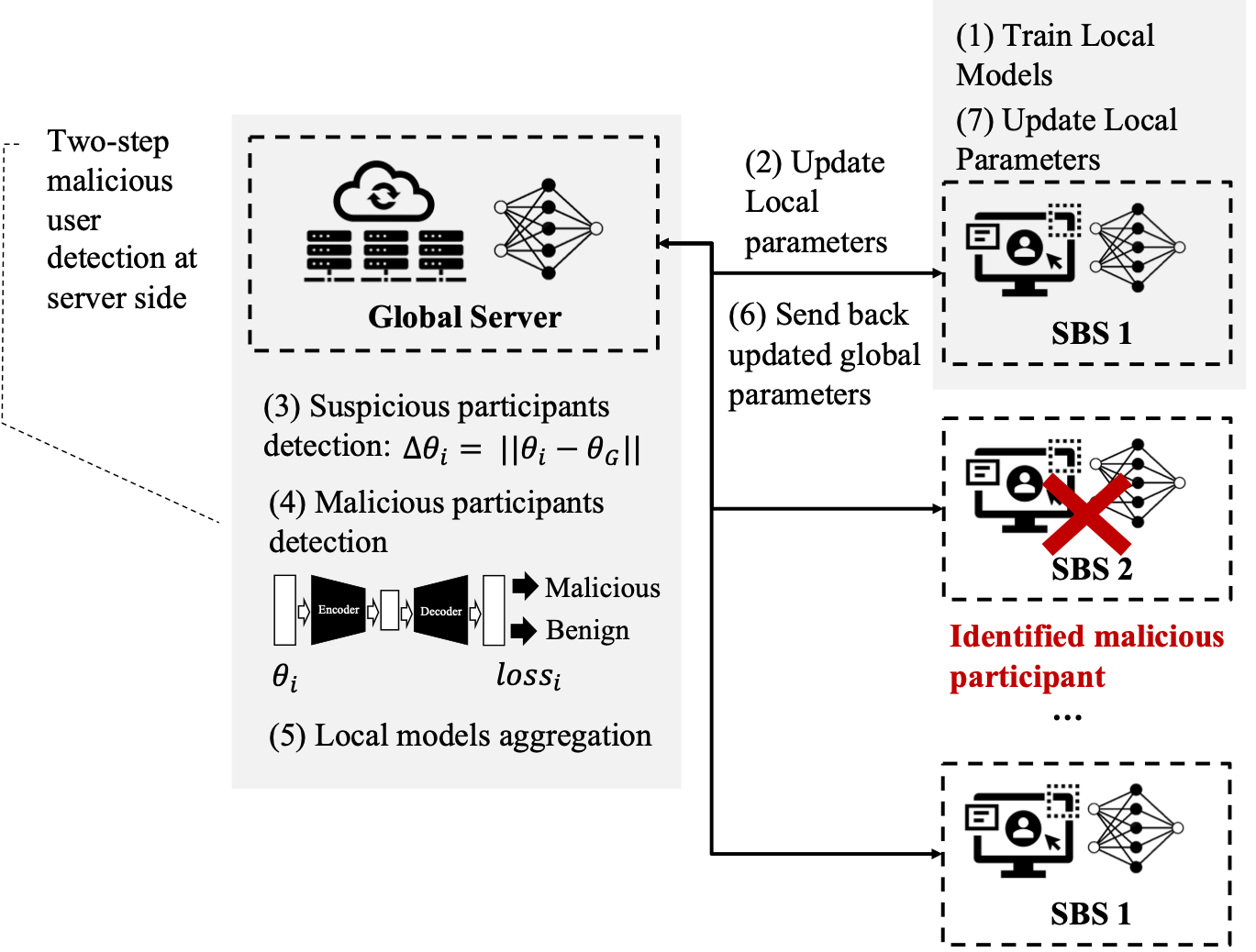}
\caption{The procedure of autoencoder-based two-step defense.}
\label{fig3}

\end{figure}

\subsection{Autoencoder-based Two-step Defense}
The procedure of an autoencoder-based two-step defense scheme is shown in Fig. \ref{fig3}. The core idea is to make the global server find the malicious participants according to the parameters of all the local models and only aggregate parameters of benign local models during the global aggregation. 

To identify the malicious models from the benign ones, we perform a two-step malicious user detection on the global server side. In the first step, a coarse detection is performed to make a fast decision on the reliable participant list. After receiving local model parameters, the global server will calculate the distance of parameters of each local model from the previous update, which can be calculated as:
\begin{equation}
G^{t+1}_n = \left\|\theta^{t+1}_n-\theta^{t}_{Global}\right\|_2,\label{eq11}
\end{equation}
where $\theta^{t}_n$ denotes the parameter of the $n^{th}$ local model at the $t^{th}$ iteration.
The similarity between the two local models can then be measured as
\begin{equation}
D^{t+1}_{nk} = \left\|G^{t+1}_n-G^{t+1}_k\right\|_2.\label{eq12}
\end{equation}

After that, we calculate the average distance of each local model from the other local models:
\begin{equation}
D^{t+1}_{n} = \frac{1}{N}\sum_{k=1}^N D^{t+1}_{nk}.\label{eq13}
\end{equation}

Then we sort all the local models according to their average distance from the other local models, and only a few local models with the smallest average distance are selected as a reliable local model set.

In the second step, an autoencoder is used to make a more accurate decision about the benign participant list. This is based on the hypothesis that if a data pattern is used in training, the model would be expected to have a lower loss than a data pattern not present in training datasets \cite{yeom2018privacy}. By using an autoencoder, we compare the differences between local model parameter data patterns instead of directly comparing the distances between local model parameters. This can reduce some of the impact of data heterogeneity on the defense.

First, an autoencoder is trained by the parameters of models in the reliable local model set. The autoencoder is then used to reconstruct the parameters of all the local models, and the reconstruction error can be formulated as:
\begin{equation}
E_i = ||\theta_{local}^i-f_2(f_1(\theta_n))||_2,\label{eq14}
\end{equation}
where $f_1$ and $f_2$ denote the encoder and the decoder. If the reconstruction error of a local model is larger than the average reconstruction error of all the local models, then the local model will be identified as the malicious local model and will not be included in the global model aggregation. In this way, the global server can prevent the compromised SBSs from poisoning the global model and other benign SBSs.

\begin{algorithm}[t]
\caption{Knowledge distillation-based distributed defense}\label{alg:KD}
\begin{algorithmic}[1]
    \Initialize{$\theta_{Local}$ of each local model, $s^{t-1}$ and $a^t$}
    \While {$TTI < T^{total}$}
        \For{each SBS}
            \State Observe local state.
            \State $S^{t}_n \gets \{\delta_n, L_n, L_0, d_n, b_n\}$.
            \State Get immediate reward $r^t_n$.
            \State Record experience $\{s^{t-1}_n,s^{t}_n,a^t_n,r^t_n\}$.
            \State Sample a mini-batch from experience buffer
            \State Calculate output Q-values of given training sample through local model and meme model according to Equation (\ref{eq_KD1}) and Equation (\ref{eq_KD2})
            \State Calculate KL divergence according to Equation (\ref{kl1}) and Equation (\ref{kl2})
            \If{$KL(p^{M}||p^{L}) < \Theta$}
                \State Train local model with Equation (\ref{eq_KD4})
            \Else
                \State Train local model with local training samples
                \State Train meme model with Equation (\ref{eq_KD3})
            \EndIf
            \State Choose actions by $a^t_n \gets \underset{a^t_n}{max}Q(s,a;\theta_{Local})$.
            \State Execute action $a^t_n$. 
            Submit $\theta_{Meme}$ to the global server.
        \EndFor
        \State Update global model according to Equation (\ref{eq_ag})
        \For{each UE} 
            \State Update local models $\theta_{Meme}=\theta_{Global}$
        \EndFor
        \State $TTI \gets TTI + 1$
    \EndWhile
\end{algorithmic}
\end{algorithm}

\subsection{Knowledge Distillation-based Defense}
Although identifying malicious participants according to the similarity of model parameters is a good defense against many attacks, it may lose effectiveness when the attacker purposely designs stealthy attacks \cite{bhagoji2019analyzing}. Therefore, we proposed another KD-based defense to defend against attacks on the client sides.
\begin{figure*}[!t]
\centering
\includegraphics[width=6.5in]{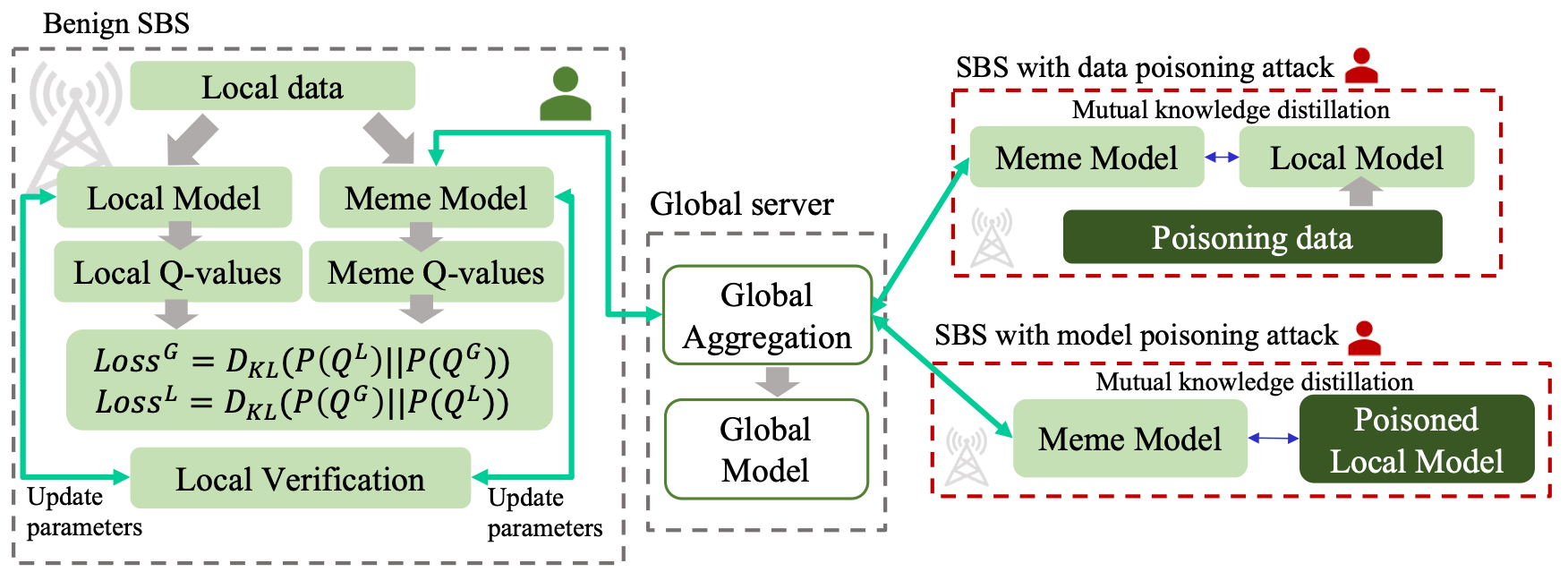}
\caption{The structure of KD-based defense against malicious participants in FL.}
\label{fig4}

\end{figure*}

The procedure of the proposed defense scheme is shown in Fig. \ref{fig4}. With this defense mechanism, we can provide dual protection for local models. Instead of directly aggregating local models, we set up a meme model for each SBS as a copy of the global model and perform mutual KD between the local model and the meme model. During the mutual KD process, models deliver only vital knowledge, and poisoning factors may be filtered out. This is the first level of protection in this scheme. Other than that, a local verification step is set up for each SBS to determine the direction of knowledge flow between the meme model and the local model. It can be performed by measuring the similarity between the local and meme model outputs based on Kullback–Leibler (KL) divergence. This is the second level of protection.

During each iteration, the updated parameters of the global model are copied to the meme models of all the SBS. The local training data is then fed to the local and meme models to generate local Q and meme-Q values. The KL divergence between the policies held by the local and meme models can be calculated by converting the Q-values into probability distribution with softmax function\cite{rusu2015policy}. It can be formulated as:
\begin{equation}
D_{KL}(p^{L}||p^{M}) = \sum_{i \in a} softmax(q^L_i)\ln\frac{softmax(q^L_i)}{softmax(q^M_i)},\label{kl1}
\end{equation} 
\begin{equation}
D_{KL}(p^{M}||p^{L}) = \sum_{i \in a} softmax(q^M_i)\ln\frac{softmax(q^M_i)}{softmax(q^L_i)},\label{kl2}
\end{equation}
where $\textbf{a}$ is the action space. $q^L_i$ denotes the output q values of the local model. $q^M_i$ denotes the output $q$ values of the meme model. It can be given as:
\begin{equation}
\textbf{q}^L = Q(s,a;\theta_{Local}),\label{eq_KD1}
\end{equation} 
\begin{equation}
\textbf{q}^M = Q(s,a;\theta_{Meme}).\label{eq_KD2}
\end{equation}

To prevent malicious local models from poisoning benign models through global model aggregation, we perform a local verification by comparing the KL divergence value with a threshold value. If $KL(p^{M}||p^{L}) > \Theta$, the global model is supposed to be quite different from the local model. This means that based on the validation results of the local data, the global model may have already been poisoned. Therefore, we only perform a one-way KD from the local model to the global model and stop KD from the global model to the local model. This can be implemented by using the KL divergence as the loss function of the meme model, which can be formulated as:
\begin{equation}
L(\theta_{Meme}) = D_{KL}(p^{L}||p^{M};\theta_{Meme}),\label{eq_KD3}
\end{equation} 
while the local model will keep using the conventional DQN loss function, which is only related to local data and not to KL divergence.

On the other hand, if $KL(p^{M}||p^{L}) < \Theta$, the local model will distill knowledge from the global model by adding the KL divergence to the loss function of the local model, which can be formulated as:
\begin{align}
L(\theta_{local}^t) = \xi (r^t+\gamma \max_a Q(s^{t+1},a;\theta_{Local}^t)\label{eq_KD4}\\-Q(s^t,a^t;\theta_{Local}^t))^2 + (1-\xi)D_{KL}(p^{M}||p^{L};\theta_{Local}^t),\nonumber
\end{align}
where $\xi$ is a parameter used to balance the weights between global and local knowledge. At the same time, the KD from the local model to the global model is stopped. In this way, the global model will avoid learning similar knowledge repeatedly and help to learn more knowledge that is different from the current knowledge. The KD-based defense scheme is summarized in Algorithm \ref{alg:KD}.

By setting up the local verification scheme, it is guaranteed that there is an upper limit to the effectiveness of the attacks on the benign models involved in FL. This upper limit can be quantified regarding the probability of the model choosing a non-optimal action. The proof is given as follows.

$Preposition\ 1$:
The upper bound of the probability of a benign local model choosing a non-optimal action under any type of attack with the proposed defense scheme is:
\begin{equation}
P_{NO} = \frac{\Theta}{\overline{E_0}},
\end{equation}
where:
\begin{equation}
\begin{split}
E_0 = -(e^{q_1}+e^{q_2})\ln(e^{q_1}+e^{q_2})+ \\e^{q_1}\ln(2e^{q_1})+e^{q_2}\ln(2e^{q_2}).
\end{split}
\end{equation}

$Proof$:
To quantify the upper bound, we first assume that in the absence of an attack, the local data can always fully train the local model to output the best action. The sleep control model has three actions, denoted as $i=1,2,3$. According to Eq. (\ref{kl2}), we have the calculation of the KL divergence on a given batch training data $X$ as follows:

\begin{equation}
    D_{KL}(P^{L}||p^{M}) = \sum_{x\in X}\sum_{i=1,2,3} softmax(q_i^{L})\ln\frac{softmax(q_i^{L})}{softmax(q_i^{M})}).
\end{equation}

By performing the local verification, the KL divergence of the local model and meme model is smaller than the threshold $\Theta$, which can be given as:
\begin{equation}
    D_{KL}(P^{L}||p^{M})\leq \Theta|X|.\label{eq_thresh}
\end{equation}

Then we suppose there exists a MDP record $x_0 = {S_0, a_0, r_0}$ in the local training batch $X$, and the outputs of the local DQN model under $s_0$ with the largest, the second largest, and the smallest values are $q_1>q_2>q_3$. The output of meme DQN model under $s_0$ is $q'_1,q'_2,q'_3$. Then we have:
\begin{equation}
    softmax(q_i^{L}) = \frac{e^{q_i}}{\sum_{j = 1,2,3}e^{q_j}}, 
\end{equation}
\begin{equation}
    softmax(q_i^{M}) = \frac{e^{q'_i}}{\sum_{j = 1,2,3}e^{q'_j}}.
\end{equation}

To make an effective attack under state $s_0$, the meme model is supposed to choose different actions from the local model. At the same time, the attacker wants to cause as many attacks as possible within the limitation of the threshold, so it may expect a minimal KL divergence when performing an attack. We also suppose an ideal situation for the attacker, where the meme model can output any distribution the attacker wants. So we can formulate the problem as:
\begin{equation}
    \begin{split}
    \min_{q'_1, q'_2, q'_3}&D^{x_0}_{KL}(P^{L}||p^{M})\\
    s.t.\ \ &q'_2>q'_1\ \ or\ \ q'_3>q'_1.\label{eqp1}
    \end{split}
\end{equation}
% $Preposition\ 1$: The optimal solution of Eq. (\ref{eqp1}) is:
% \begin{align}
%     q'_1 = q'_2 &= \frac{e^{q_1}+e^{q_2}}{2},\\
%     q'_3 &= q_3,\nonumber
% \end{align}
% $Proof$:
To solve this problem, the KL divergence can be firstly expanded as:
\begin{equation}
    \begin{split}
    D^{x_0}_{KL}(P^{L}||p^{M}) = &\sum_{i=1,2,3}[\frac{e^{q_i}}{\sum_{j = 1,2,3}e^{q_j}}(q_i-\ln(\sum_{j = 1,2,3}e^{q_j})\\
    &-q'_i+ln(\sum_{j = 1,2,3}e^{q'_j}))].
    \end{split}
\end{equation}

Since $q_1,q_2,q_3$ are fixed values, irrelevant items can be removed from the expression of KL divergence and simplify the optimization problem as:
\begin{equation}
    \begin{split}
    \min_{q'_1, q'_2, q'_3}&\sum_{i=1,2,3}[e^{q_i}\ln(\sum_{j = 1,2,3}e^{q'_j})-q'_ie^{q_i}]\\
    s.t.\ \ &q'_2>q'_1\ \ or\ \ q'_3>q'_1.
    \end{split}
\end{equation}

According to the definition of KL divergence, the minimum value is achieved when the output of two models have the same distribution, which is $q'_1,q'_2,q'_3 = q_1,q_2,q_3$. Since the optimization function is smoothly continuous, the value of the function will increase in any direction near the minimum value point. Since we also have the prerequisite $q_1>q_2>q_3$, the solution of the given optimization problem lies on the tipping point that makes $q'_1 = q'_2$ and $q'_3 = q_3$. Supposing $q'_1 = q'_2 = q^*$,
\begin{equation}
    \begin{split}
    \hat{D^{x_0}}_{KL}(P^{L}||p^{M}) = \sum_{i=1,2,3}[e^{q_i}\ln(2e^{q^*}+e^{q^3})]\\-q^*(e^{q_1}+e^{q_2}),
    \end{split}
\end{equation}

where $\hat D$ denotes the expression of KL after simplification.
The derivation of $\hat{D^{x_0}}_{KL}(P^{L}||p^{M})$ can then be given as:
\begin{equation}
    \begin{split}
    \frac{\partial{\hat{D^{x_0}}_{KL}(P^{L}||p^{M})}}{\partial{q^*}} = 2e^{q^*}\frac{\sum_{i=1,2,3}e^{q_i}}{2e^{q^*}+e^{q_3}}-e^{q_1}-e^{q_2}.
    \end{split}
\end{equation}

When the derivation is zero, the minimal value point of $q^*$ can be calculated as:
\begin{equation}
    \begin{split}
    q^* = \frac{e^{q_1}+e^{q_2}}{2}.
    \end{split}
\end{equation}

The minimal KL divergence to make an effective attack, which is denoted as $E_0$, can be calculated as:
\begin{equation}
\begin{split}
E_0 = -(e^{q_1}+e^{q_2})\ln(e^{q_1}+e^{q_2})+ \\e^{q_1}\ln(2e^{q_1})+e^{q_2}\ln(2e^{q_2}).
\end{split}
\end{equation}

We suppose there is sufficient local training data that is uniformly distributed. The average minimal KL divergence to make an effective attack can be estimated by calculating the average $E_0$ over the batch training data $X$, which can be given as:
\begin{equation}
\begin{split}
\overline{E_0} = -\frac{\sum_x\in X(e^{q_1}+e^{q_2})\ln(e^{q_1}+e^{q_2})}{|X|} \\ + \frac{\sum_x\in X e^{q_1}\ln(2e^{q_1})+e^{q_2}\ln(2e^{q_2})}{|X|}.
\end{split}
\end{equation}
According to Eq. (\ref{eq_thresh}), there is an upper bound for the KL divergence. That means the number of action choices that an attacker can change is limited, or the model will choose to learn locally without being affected by the attacks. Therefore, after enough training rounds, the upper bound of the probability of a benign local model choosing a non-optimal action can be estimated as:
\begin{equation}
P_{NO} = \frac{\Theta}{\overline{E_0}}.
\end{equation}

This concludes the proof.

We can give an intuitive interpretation of this result. $E_0$ in the denominator part represents the average energy required for the model to be attacked and indicates the degree of vulnerability of the model to attack. The numerator is the threshold for constraining the dispersion, which regulates how closely the models cooperate in FL. These two factors determine the extent to which the benign models are attacked under the given defense mechanism in FL.

According to this proof, we demonstrate that the performance of our proposed KD-based defense method is not limited by the attack methods and can theoretically be effective against any type of poisoning attack.

\vspace{-10pt}
\subsection{Baseline Defense: Krum}
To make a comparison with two proposed defense schemes, we use a Krum defense proposed in \cite{blanchard2017machine} as a baseline. In the Krum defense, the global server will calculate the average distance of each local model from the other local models according to Eq. (\ref{eq13}) and choose the local model with the smallest average distance. The parameters of the global will be replaced by the parameters of the chosen local model in each iteration.

\vspace{-10pt}
\subsection{Computational complexity and overhead analysis}

In this subsection, we analyze the computational complexity and the overhead of two proposed defense methods. In the autoencoder-based two-step defense method, the main computation comes from calculating the model similarity and training the autoencoder. The model similarity is measured by the Euclidean distance, and the computational complexity is proportional to the dimension of model parameters $len(\theta)$. The complexity of the autoencoder model is proportional to the model layers and the number of neurons. The computational complexity of the autoencoder-based two-step defense method can be formulated as $O(len(\theta)+\sum_{i=1}^{I_A}N^A_{i-1}N^A_{i})$, where $N^A_{i}$ denotes the number of neurons of the autoencoder at the $i^{th}$ layer. $I_A$ denotes the number of layers of the autoencoder.

In the KD-based defense method, the main computation comes from the inference of the meme model and the calculation of the KL divergence. The computational complexity of meme model inference is proportional to the model layers and the number of neurons of the meme model. The computational complexity of KL divergence calculation is proportional to the size of the action space. So the computational complexity of the KD-based defense method is $O(len(a)+\sum_{i=1}^{I_M}N^M_{i-1}N^M_{i})$, where $N^M_{i}$ denotes the number of neurons of the meme model and $len(a)$ denote the size of the action space.

Both of the defense methods do not introduce additional overhead to the system. In the KD-based defense method, instead of submitting the local model parameters to the global server, the meme model parameters are submitted. However, in this paper, since the two models are of the same size, the overhead remains unchanged.

\vspace{-10pt}
\subsection{Attackers and defenders analysis}
In this section, we compare and analyze the knowledge of the attackers and the defenders to carry out different attacks and defenses. The following analysis demonstrates the feasibility of the proposed attack and defense methods.

For GAN-enhanced attacks, the only knowledge attackers have is the structure and parameters of the global model. They do not have any training data. This kind of knowledge can be obtained by intercepting the over-the-air transmission channel between the global server and the clients of FL. Specifically, like Sybil attacks on distributed systems, the attacker can inject fake participants into the FL system instead of controlling existing participants \cite{cao2022mpaf}.

For regularization-based model poisoning attacks, the attackers are assumed to know the local training data of the malicious participants and the structure and parameters of the global model. They do not have access to the local training data of other benign participants.

To perform autoencoder-based defense and KD-based defense, the global server has access to the local model parameters updated by each participant. It does not have access to the local datasets or other information about the participants. The participants have access to their local data and the shared global parameters. They do not have access to the data of other participants.

\vspace{-10pt}
\section{Numeric Results}
\subsection{Simulation Settings}
To evaluate the effect of attacks and defenses under the wireless communication system, we first simulated a near-realistic 3GPP urban macro network environment and built the channel model and communication model with Python 3.8. Next, we modeled the data transmission between the SBS and the UE with the simulation platform, and in this process, MDP data was generated and collected at each SBS. These data are used as the training data for each local DQN model. In each transmission time interval (TTI), a set of MDP data is collected by each SBS. During the simulation, we consider a heterogeneous network with 1 MBS and 20 SBSs. To configure a non-IID data distribution among local models, we set various numbers of UEs and different traffic loads for each SBS. The propagation path loss is set as $128.1+37.6log(Distance)$. The simulations are repeated in ten runs with a 95\% confidence interval. The coefficients of rewards $\omega_1$, $\omega_2$, and $\omega_3$ are 0.25, 1, and 0.05. Other simulation details are given in Table \ref{table1}. In each episode, we simulate a 24-hour typical residential area traffic pattern
which is given from \cite{zhou2022hierarchical}.

\begin{table}[t]
\caption{Simulation settings.}
\renewcommand{\arraystretch}{1.2}
\centering
\begin{adjustbox}{width=\columnwidth,center}
\begin{tabular}{|l|l|}
\hline
\textbf{Wireless network settings}      & \textbf{Traffic load settings}             \\ \hline
3GPP urban macro network                & Peak traffic load of each SBS:             \\
Bandwidth: 20MHz                        & 12/14/16/18 Mbps                        \\ \cline{2-2} 
Number of RBs: 100                      & \textbf{Sleep mode settings}               \\ \cline{2-2} 
Subcarriers in each RB: 12              & Energy consumption of active mode: 100\%    \\
Subcarrier bandwidth: 15kHz             & Energy consumption of sleep mode: 50\%      \\ \cline{1-1}
\textbf{BS and UE settings}             & Energy consumption of deep sleep mode: 15\% \\ \hline
Number of SBS: 20                    & \textbf{DQN settings}                      \\ \cline{2-2} 
Transmission power of MBS: 40W       & Initial learning rate: 0.01                \\
Transmission power of SBSs: 4W          & Discount factor: 0.8                       \\
Carrier frequency of SBS: 3.5 GHz       & Epsilon value: 0.05                        \\
Carrier frequency of MBS: 2.0 GHz       & Hidden layer: 2                            \\
Number of UEs: 3-11 each SBS            & Batch size: 256                            \\ \hline
\end{tabular}
\end{adjustbox}
\label{table1}
\vspace{-15pt}
\end{table}

According to the MDP definitions, the dimension of the state and the next state is 16, and the dimension of the action and the reward is 1. Therefore, the dimension of the MDP data is 34. The input dimension of the DQN model is 16 and the output dimension is 3. The DQN has two layers, the first layer has 64 neurons and the second layer has 32 neurons.

We use two communication metrics to evaluate the results, the system throughput and the energy efficiency. The purpose of attacks is to reduce system throughput and energy efficiency, while the purpose of defense methods is to recover system throughput and energy efficiency to the values that it would have without the attacks. To better evaluate our proposed defense methods, we use a well-known Krum defense method proposed in \cite{blanchard2017machine} as a baseline for comparison.

\vspace{-10pt}
\subsection{Model distribution analyses}
\begin{figure}[t]
    \centering
    \subfigure[The distribution of local models with IID data.]{
        \includegraphics[width=7.5cm,height=5cm]{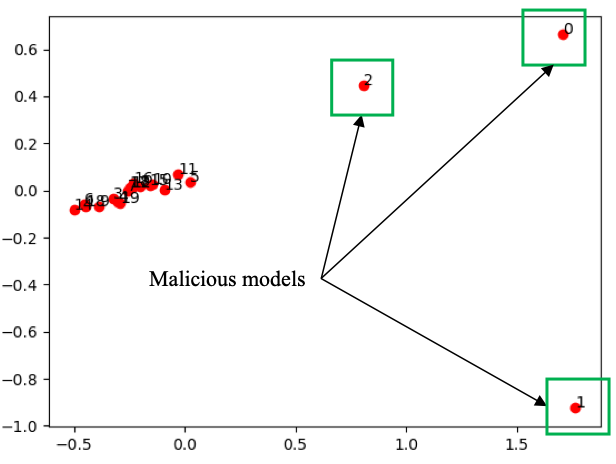}
    }
    \subfigure[The distribution of local models with non-IID data.]{
	\includegraphics[width=7.5cm,height=5cm]{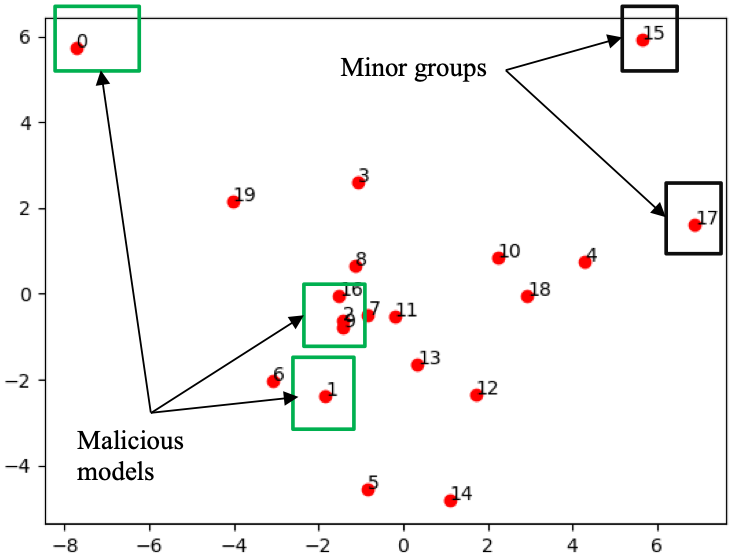}
    }
    \caption{A comparison between IID and non-IID data.}
    \label{fig5}
    \vspace{-20pt}
\end{figure}

This section presents the model distribution analyses under FL with IID data and non-IID data. To generate IID data, we set identical traffic loads, UE numbers, and UE distribution of each SBS. In this way, the distributions of local models will be very close to each other. However, this is an ideal situation that is almost impossible to happen in practice. For non-IID data, we set different traffic loads, UE numbers, and UE distribution of each SBS. So distributions of local models will be far from each other. To visualize the high-dimensional model parameters data, we first perform the principal component analysis (PCA) to extract two principle components and project them into a two-dimensional coordinate system. Suppose we have $m$ participants and each participant has a DQN model with $n$ parameters. We consider it as a set of n-dimensional data with $m$ samples. During the PCA, the samples are first centralized. Then, the covariance matrix for all samples is calculated, and the eigenvalue decomposition of the matrix is performed. The eigenvectors corresponding to the two largest eigenvalues are taken out, and all the eigenvectors are normalized to form an eigenvector matrix. This matrix can transform the high-dimensional model parameter samples into two-dimensional points as the PCA result. The projection result after PCA is shown in Fig. \ref{fig5}. In this figure, the $x$ and $y$ axes are the values of two principal components of local model parameters after PCA.

Fig. \ref{fig5} (a) shows the distribution of local models with IID data, and Fig. \ref{fig5} (b) shows the distribution of local models with non-IID data. We also perform a data poisoning attack on the first three local models. It can be observed that in Fig. \ref{fig5} (a), benign models are very close to each other, and malicious models are far away from them. This illustrates that with IID data, malicious models can be easily identified. However, in Fig. \ref{fig5} (b), benign models are not as close as the situation with IID data, and it is tough to intuitively identify the malicious models and benign models from model distribution. And benign models with heterogeneous data distribution from the major data, called minor groups, may be identified as malicious models. This demonstrates our motivation for starting this research that the attack and defense problems of FL with non-IID data are more complex and worth exploring than those of FL with IID data.

\vspace{-10pt}
\subsection{Performance of Attacks}

\begin{figure}[t]
    \centering
    \subfigure[The system throughput under different attacks.]{
        \includegraphics[width=7.5cm]{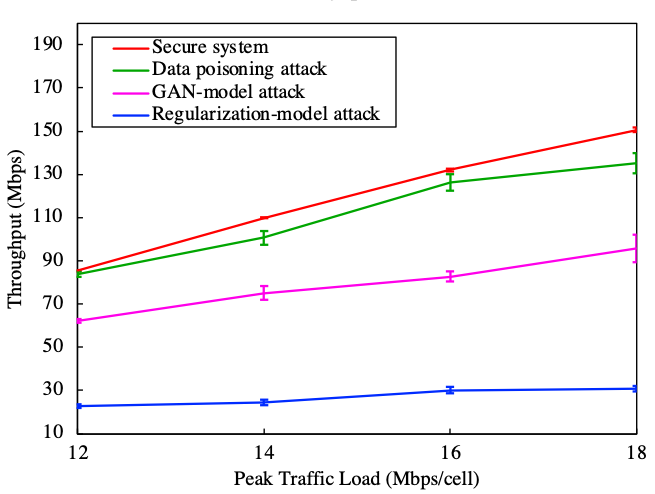}
    }
    \subfigure[The energy efficiency under different attacks.]{
	\includegraphics[width=7.5cm]{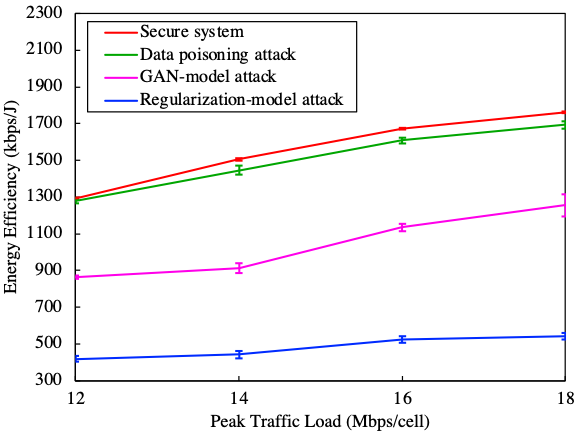}
    }
    \caption{The system performance under three different attacks.}
    \label{fig7}
    \vspace{-15pt}
\end{figure}

\begin{figure}[ht]
\centering
\includegraphics[width=9.2cm]{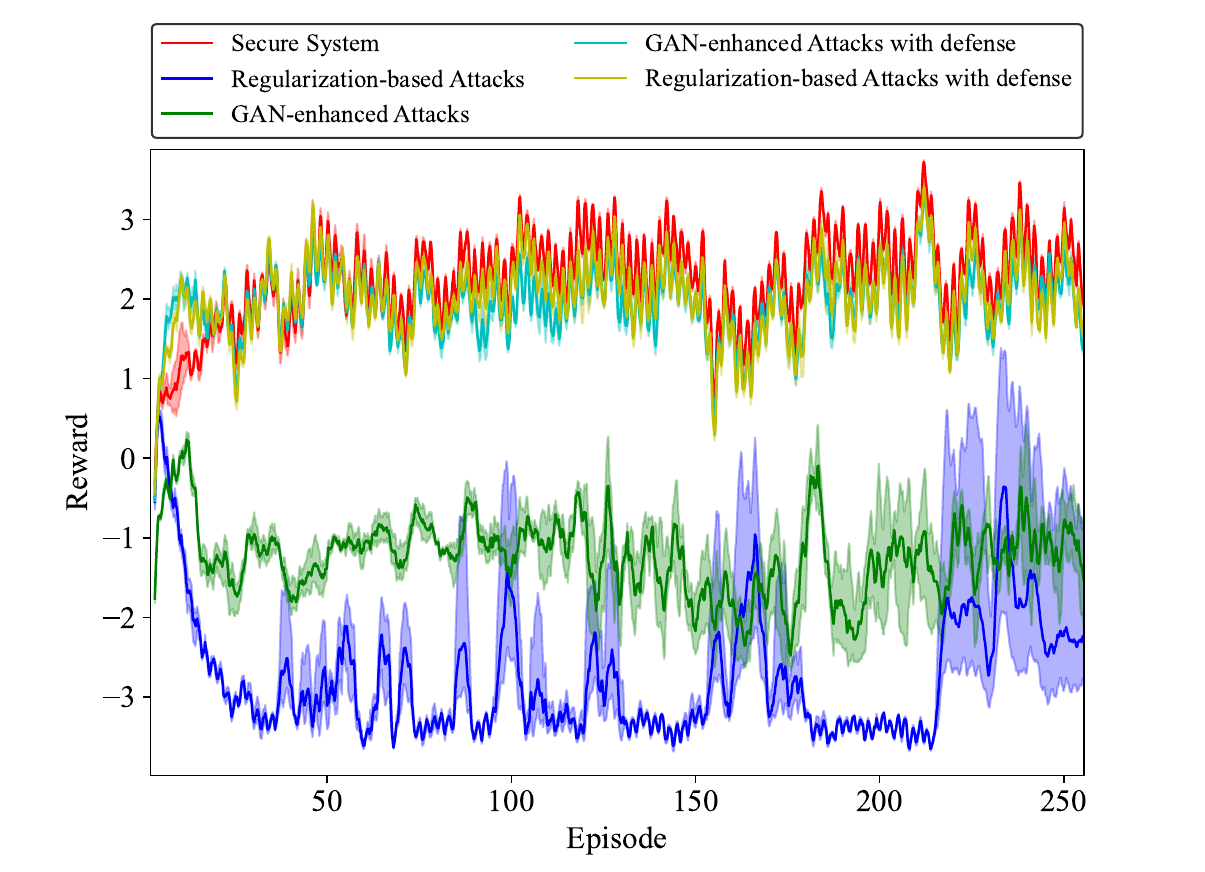}
\caption{The convergence curves of the system reward without attacks, under GAN-based attacks, under regularization-based attacks, and proposed defense.}
\label{fig7-1}
\vspace{-10pt}
\end{figure}

Fig \ref{fig7} shows the average downlink throughput of all the UEs and energy efficiency under two proposed intelligent attacks, GAN-based model poisoning attack and regularization-based model poisoning attack, and one baseline, data poisoning attack, when there are three attackers out of twenty participants. We also simulated with more attackers, and the results confirmed our findings. However, the number of attackers will be low in practice due to the cost of implementing attacks. The energy efficiency is the ratio of the transmission data of all UEs to the overall energy consumption of the system, which is defined in Eq. \eqref{eq3}. We also add the performance of the secure system for comparison. During the simulations, we change the peak traffic load of the SBS from 12 Mbps to 18 Mbps. 
From these figures, it can be concluded that among these attacks, model poisoning attacks are more effective than data poisoning attacks. In the proposed data poisoning attack, we changed about 5\% of the MDP data. With three attackers, it only decreases the system throughput by 10.2\% and reduces the energy efficiency by 3.9\%. In contrast, the other two proposed model poisoning attacks have more substantial effects on the system performance. For the GAN-enhanced model poisoning attack, it decreases the system throughput by 36.5\% and the energy efficiency by 28.7\%. For the regularization-based model poisoning attack, it reduces the system throughput by 69.3\% and the energy efficiency by 79.5\%.

The reason why model poisoning attacks are more effective is that such attacks can directly change the parameters of the model, and the influence is more definitive. The data poisoning attacks, however, only take effect when poisoning data is selected for training in a local training iteration. Besides, the unpolluted data also places constraints on the model parameters when the model is updated so that the poisoned data do not necessarily influence the model parameters. 

The effectiveness of data poisoning will indeed increase as the percentage of poisoned MDP data increases. However, this also means the malicious participants are easier to identify by defense schemes, which will be discussed in the following subsection. Therefore, it does not make sense to enhance the attack simply by increasing the amount of poisoned data. A more practical way to perform intelligent attacks is to design intelligent attacks like GAN-enhance model poisoning attacks and regularization-based model poisoning attacks.

Fig Fig \ref{fig7-1} shows the convergence curves of the system reward without attacks, under attacks, and under attacks and defense. We first compare the reward curve of the secure system and the curves of two different attacks. As can be observed in the figure, for the reward of FL without attacks, although there were some fluctuations due to the dynamics of the environment, the reward generally grows and eventually converges. In contrast, in the presence of attacks, the reward value decreases gradually over the process of training.

The regularization-based attack leads to a lower reward compared with GAN-enhanced attacks. This is because the regulation-based attack maximizes the loss function during training, which makes it a more proactive and aggressive attack. In contrast, the GAN-enhanced attack is carried out by generating synthetic fake model parameters. These fake model parameters do not contain specific meanings such as increasing the system loss, but only serve as a disturbance to the global model. So the GAN-based attack is not as aggressive as the regularization-based attack. Also, a noticeable oscillation in the reward values of the regularization-based attack is shown. This is because the loss function of the regularization-based attack model consists of two components, an attack component and a regularization component. These two components act jointly and cause the system reward value to be unstable. 

Moreover, Fig \ref{fig7-1} also incorporates our proposed KD-based defense method for each of the two attacks, and the reward convergence curves after defense are shown. As it can be observed, after the defense, the system reward is close to the reward of the secure system, which illustrates the effectiveness of our proposed defense method. In the next subsection, we provide more simulation results and an analysis of our proposed defense methods.

\vspace{-10pt}
\subsection{Performance of Defense Methods}
This section investigates the performance of two proposed defense schemes, autoencoder-based defense and KD-based defense, and one baseline defense scheme, Krum-based defense, under three attacks mentioned in the last subsection. The two model poisoning attacks we proposed can be regarded as upgraded versions of existing poisoning attacks. According to the comparison with the data poisoning attack baseline, they are more intelligent and stealthy and, therefore, more difficult to defend against. So, the three attacks we used for simulation are sufficiently representative to validate the effectiveness of our proposed defense methods. While our defense methods are capable of protecting the models from these intelligent attacks, this implies that they are also effective against other common poisoning attacks.

\begin{figure}[t]
    \centering
    \subfigure[The system throughput under different defense schemes.]{
        \includegraphics[width=7.5cm]{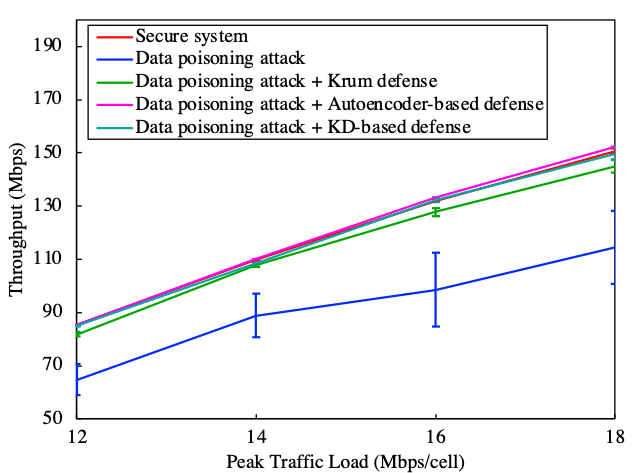}
    }
    \subfigure[The energy efficiency under different defense schemes.]{
	\includegraphics[width=7.5cm]{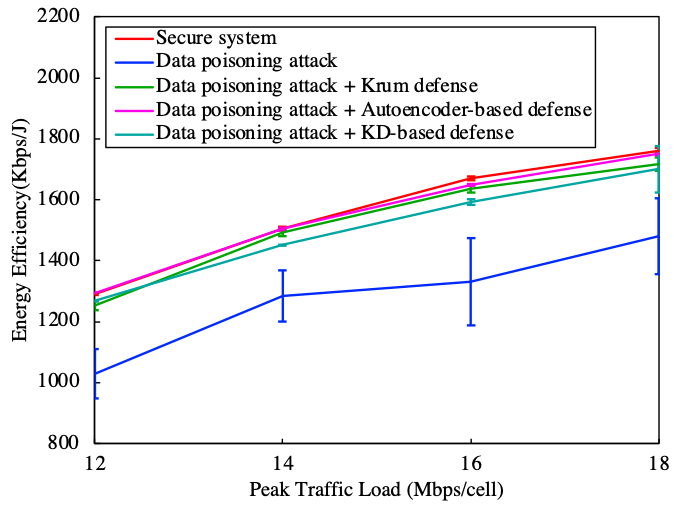}
    }
    \caption{The system performance under data poisoning attacks with different defense schemes.}
    \label{fig8}
    \vspace{-20pt}
\end{figure}

Fig. \ref{fig8} shows the system throughput and energy efficiency under data poisoning attacks with different defense schemes. We also show the performance of the secure system and the performance under attacks without any defense for comparison. The throughput and energy efficiency of the secure system are used as upper limits to compare with the performance after defense. If the throughput and energy efficiency of the system is close to the upper limit under the secure system after defense, the defense method is considered effective. In this simulation, we increased the number of attackers to five to achieve a more noticeable attack effect and elevate the difficulty of defense. However, each defense scheme can still successfully defend against the data poisoning attack. Among these three defense schemes, the autoencoder-based defense scheme has the best effect and can almost reach the upper limit of the FL system. The other two defense schemes can also improve system performance to at least 95\% of the throughput and energy efficiency in a secure system.

\begin{figure}[ht]
    \centering
    \subfigure[The system throughput under different defense schemes.]{
        \includegraphics[width=7.5cm]{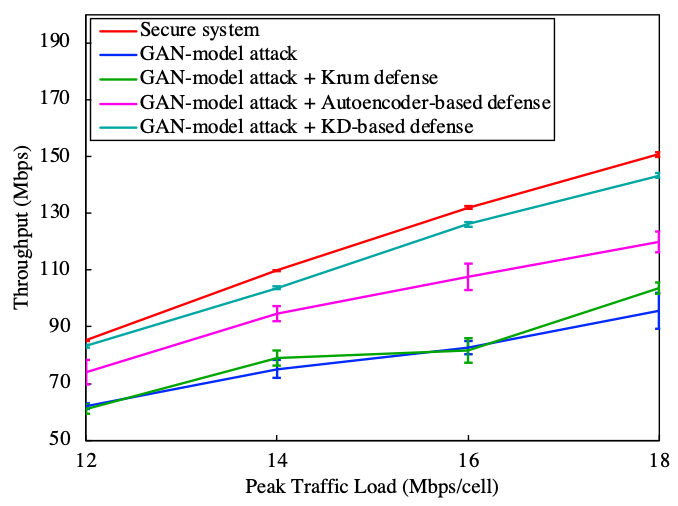}
    }
    \subfigure[The energy efficiency under different defense schemes.]{
	\includegraphics[width=7.5cm]{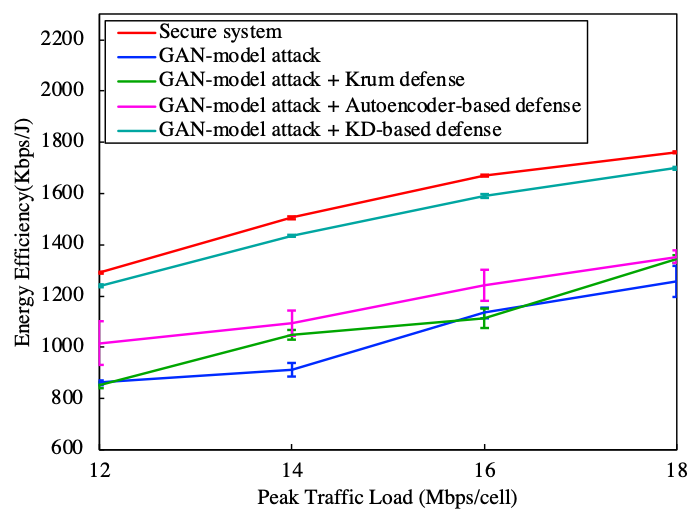}
    }
    \caption{The system performance under GAN-enhanced model attacks with different defense schemes.}
    \label{fig9}
    \vspace{-20pt}
\end{figure}

\begin{figure}[ht]
    \centering
    \subfigure[The system throughput under different defense schemes.]{
        \includegraphics[width=7.5cm]{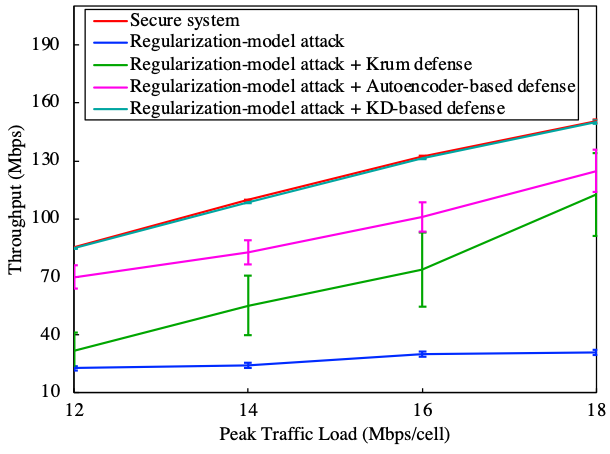}
    }
    \subfigure[The energy efficiency under different defense schemes.]{
	\includegraphics[width=7.5cm]{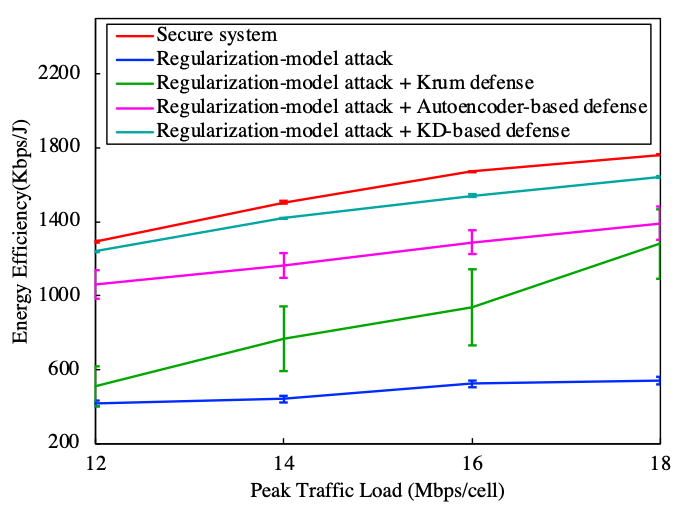}
    }
    \caption{The system performance under regularization-based model attacks with different defense schemes.}
    \label{fig10}
    \vspace{-10pt}
\end{figure}

Fig. \ref{fig9} shows the system throughput and energy efficiency under GAN-enhanced model poisoning attacks with different defense schemes. According to the simulation results discussed in the last subsection, three attackers are strong enough to bring noticeable attacks to the system performance under GAN-enhanced model poisoning attacks. So in this simulation, we set three attackers out of twenty participants. Among the three defense methods, it can be observed that Krum provides the weakest defense to the attacks. The throughput and energy efficiency after Krum defense are close to the values under attacks but without any defense scheme. 
In contrast, the effect of autoencoder-based defense is more prominent. It increases the system throughput by 29.9\% and the energy efficiency by 9.4\%. However, the system performance under this defense scheme is still much worse than a secure system. KD-based defense performs best of the three defense schemes and is closest to the upper limit of the FL system. It can improve the system performance to 95.3\% of the throughput and 95.7\% of the energy efficiency in a secure system.

Fig. \ref{fig10} shows the system throughput and energy efficiency under regularization-based model poisoning attacks with different defense schemes. It can be observed that all three defense schemes can provide defensive effect. Among the three defense methods, Krum defense performs worst. In a best-case scenario, it can only improve the system performance to 74.6\% of the throughput and 72.3\% of the energy efficiency in a secure system. Autoencoder-based defense scheme performs better and it can improve the system performance to 81.17\% of the throughput and 82.3\% of the energy efficiency in a secure system. 
Moreover, Krum defense is more favorable when the traffic load is high, while autoencoder-based defense is more advantageous when the traffic load is low. This also illustrates the instability of these two defense schemes, which means the effectiveness of the defense methods can be affected by environmental parameters. In contrast, KD-based defense is the most stable and effective defense scheme. It can increase the energy efficiency to at least 94.2\% of the performance in the secure system and achieve almost the same throughput performance as the secure system.

\begin{figure}[ht]
\centering
\includegraphics[width=7.5cm]{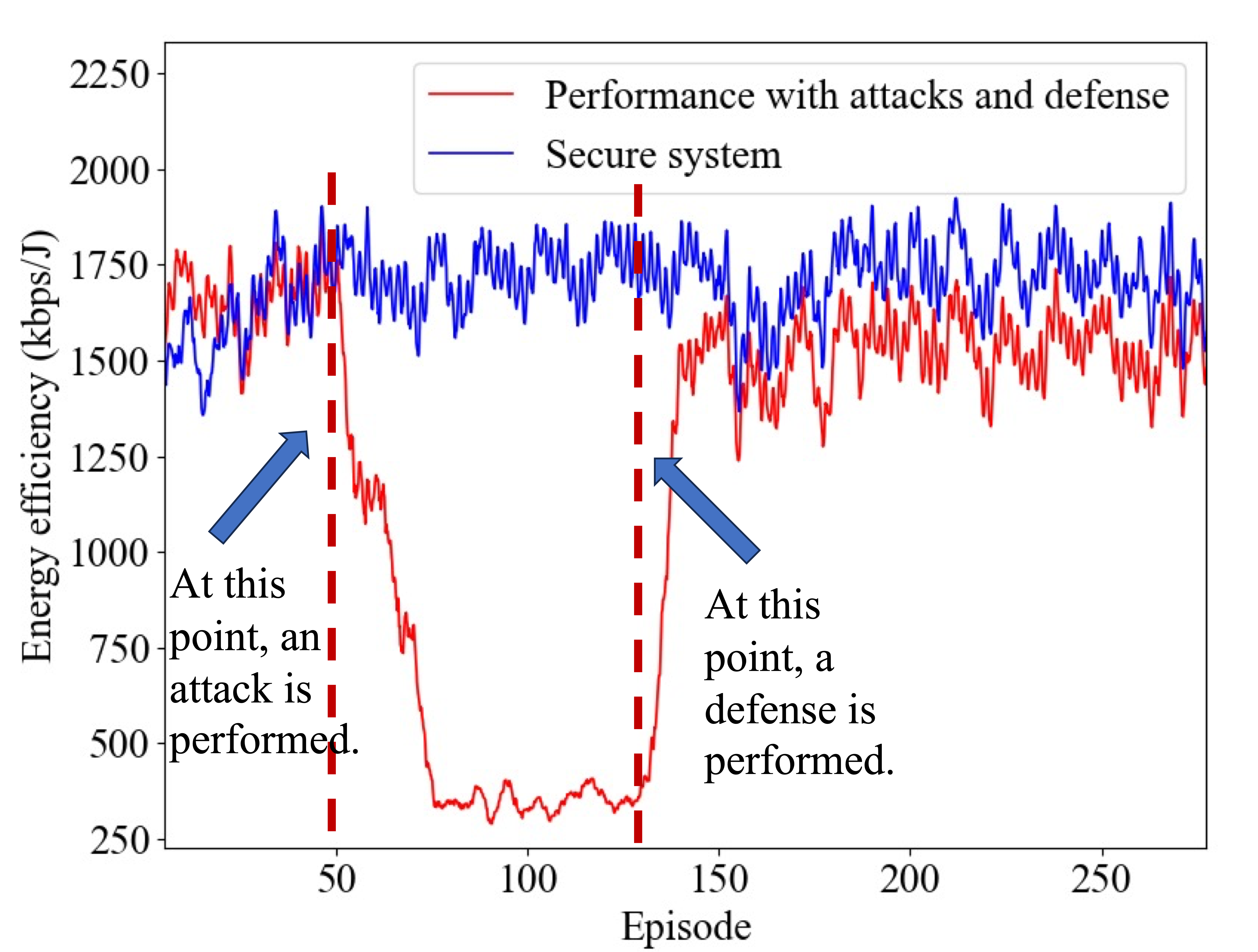}
\caption{The dynamic change of energy efficiency after attacks and defense.}
\label{fig11}
\end{figure}

\begin{figure}[ht]
\centering
\includegraphics[width=7.5cm]{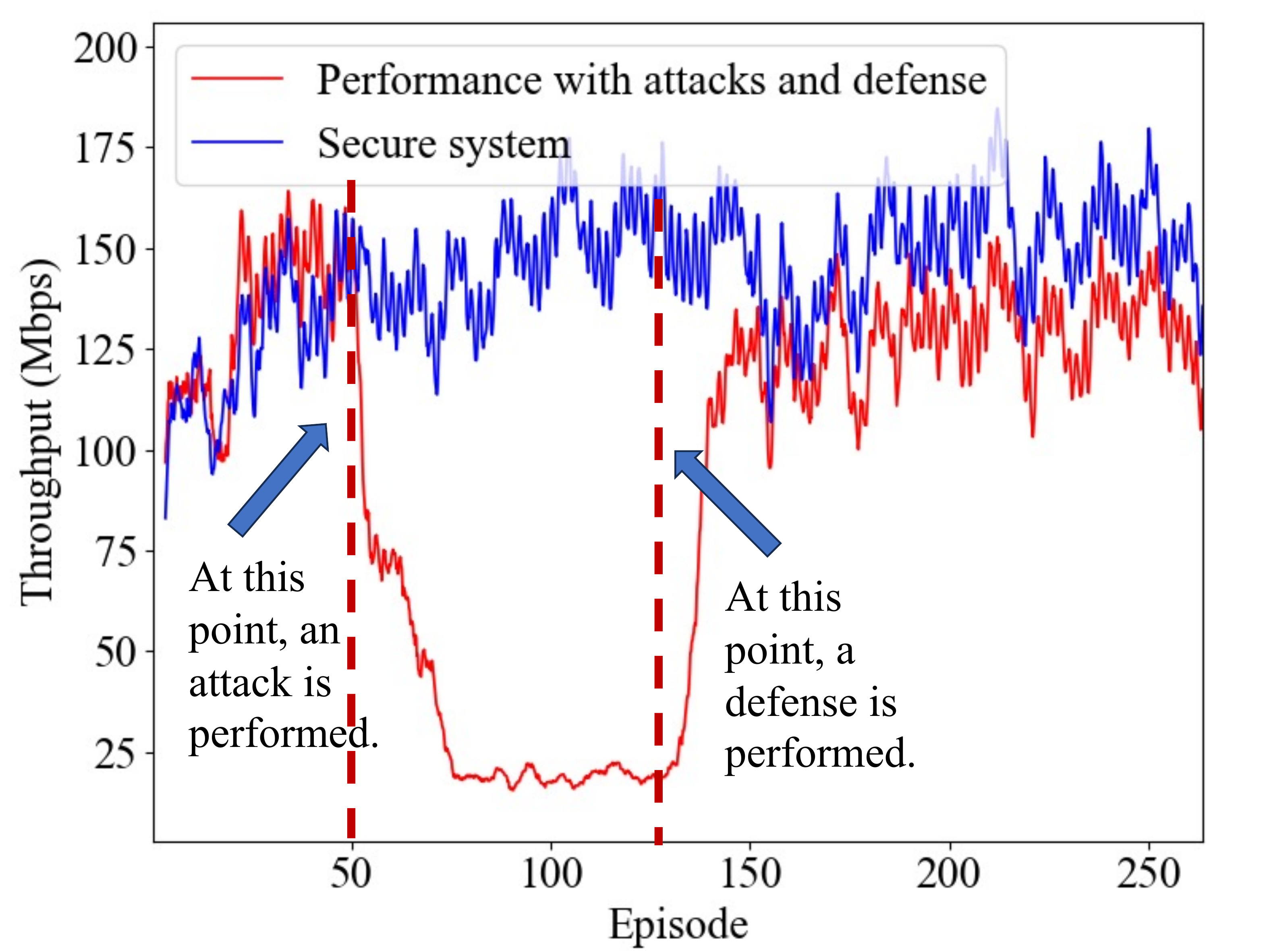}
\caption{The dynamic change of system throughput after attacks and defense.}
\label{fig12}
\end{figure}

Finally, Fig \ref{fig11} and Fig \ref{fig12} show a dynamic process of how the energy efficiency and the throughput change under intelligent attacks and defense methods. In this simulation, we take the regularization-based model poisoning attack and KD-based defense as examples. In previous simulations, both attack and defense are present throughout the simulation. Differently, in this simulation, we use a special way to conduct the simulation and add the attack and defense in the middle of the simulation to show a more dynamic attack and defense process. At about 50 episodes, we start an intelligent attack, and the attacker learns to degrade the system performance by decreasing the energy efficiency. At about 125 episodes, we start an intelligent defense scheme. The defense scheme can protect the system against attacks and increase energy efficiency.

In summary, the GAN-enhanced model poisoning attack and regularization-based model poisoning attack are more difficult to defend than the data poisoning attack. According to these results, we analyze the advantages and applicable scenarios of two proposed defense methods. First, the autoencoder-based defense method can identify the malicious participants while the KD-based defense method cannot. If the defender wants to know which participants have been attacked, the autoencoder-based defense method is the more appropriate defense method. Second, the autoencoder-based defense method performs better for simpler attack strategies where the attackers do not take any steps to make the attack more stealthy, such as the data poisoning attack \cite{tolpegin2020data}\cite{gupta2023novel}. This is because the autoencoder-based defense method detects malicious participants mainly relying on the similarity between local model parameters. Under these attacks, the local model parameters of malicious participants are largely different from benign participants and can be easily identified by autoencoders. But for more advanced attacks such as the two intelligent attacks we propose, extra strategies are added to make the local model parameters of malicious participants close to those of benign participants and to make attacks more stealthy \cite{bhagoji2019analyzing}. In this way, these malicious local model parameters sometimes cannot be easily detected by autoencoders, and the performance of the autoencoder-based defense method for these attacks is worse.

The advantage of using the autoencoder-based defense method for simpler attack strategies is that it can restore system performance to an unattacked state to a higher extent. Once the autoencoder identifies the malicious participants, the malicious local parameters are excluded from the global aggregation process. In this way, the impact of the attacks is minimized. In contrast, the KD-based defense method defends against attacks by distilling benign information. In the process of distillation, some useful information may also be lost, which makes the KD-based defense method less effective than the autoencoder-based defense method under simpler attacks. However, since the KD-based defense method does not require identifying the malicious participant, the defense effectiveness is less affected by the stealthiness of the attacker. Therefore it demonstrates better results with more advanced attacks. In addition, the KD-based defense method sets up a local validation mechanism to detect the impact of attacks on the model, and it is theoretically effective against any type of poisoning attack. We also proved that there exists an upper bound on the effectiveness of attacks on the benign models involved in FL when KD-based defense is applied. As a result, the KD-based defense method can still provide a relatively strong defense against intelligent attacks and prevent them from significantly degrading the performance of the system.

\section{Conclusion}
In this work, we study the security problem of FL for cell
sleep control application in a heterogeneous network. We consider two intelligent attacks, a GAN-enhanced model poisoning attack, and a regularization-based model poisoning attack. We also propose two defense schemes, a similarity-enabled autoencoder-based defense scheme, and a KD-based defense scheme. According to the simulation results, intelligent attacks are more effective than regular data poisoning attacks and can
degrade system performance by causing lower energy efficiency and throughput. Moreover, our proposed defense schemes can effectively defend against these attacks. Especially, the KD-based defense scheme can recover the system performance to approximately 95\% of the performance under a secure system regardless of the type of attack. For future work, we will apply the proposed methods to more public benchmarks and datasets, exploring the performance in more diverse scenarios.

\section*{Acknowledgment}
This work has been supported by MITACS and Ericsson
Canada, and NSERC Canada Research Chairs and Collaborative Research and Training
Experience Program (CREATE) under Grant 497981.

\normalem
% \begin{refcontext}[sorting = none]
% \printbibliography
\bibliographystyle{IEEEtran}
\bibliography{reference}
% \end{refcontext}

\end{document}